\documentclass{article}
\usepackage{graphicx} % Required for inserting image
\usepackage{cite}
\usepackage{amsmath,amssymb,amsfonts}
\usepackage{algorithm}
\usepackage{algpseudocode}
\usepackage{graphicx,color}
\usepackage{textcomp}
\usepackage[breaklinks=true]{hyperref}
\usepackage{breakcites}
\usepackage[caption=false, font=small]{subfig}
\usepackage{booktabs}
\usepackage[table]{xcolor}
\usepackage{multirow}
\usepackage{tabularx}
\usepackage{tablefootnote}
\usepackage{eqparbox}
\usepackage[switch]{lineno}
\algrenewcommand{\algorithmiccomment}[1]{\unskip\hfill\# #1}

\let\citet\cite
\let\citep\cite
\let\nameref\ref

\def\BibTeX{{\rm B\kern-.05em{\sc i\kern-.025em b}\kern-.08em
    T\kern-.1667em\lower.7ex\hbox{E}\kern-.125emX}}
\AtBeginDocument{\definecolor{tmlcncolor}{cmyk}{0.93,0.59,0.15,0.02}\definecolor{NavyBlue}{RGB}{0,86,125}}

\newcommand{\ie}{\textit{i.e.}, }
\newcommand{\eg}{\textit{e.g.}, }

\makeatletter
\DeclareRobustCommand{\change}{%
  \@bsphack
  \normalcolor %%% <<---- uncomment this line and comment the following two lines to go back to original black color
  \@esphack
}
\DeclareRobustCommand{\jchange}{%
  \@bsphack
  \normalcolor %%% <<---- uncomment this line and comment the following two lines to go back to original black color
  \@esphack
}
\DeclareRobustCommand{\dchange}{%
  \@bsphack
  \normalcolor %%% <<---- uncomment this line and comment the following two lines to go back to original black color
  \@esphack
}
\DeclareRobustCommand{\stopchange}{%
  \@bsphack
  \normalcolor
  \@esphack
}
\makeatother
\makeatletter
\AtBeginDocument{\let\hl\@firstofone}
\makeatother


\title{Towards Physical Underwater Robotic Assistance \\for Scuba Diver Movement in Confined Spaces}
\author{Demetrious T. Kutzke \& Junaed Sattar}
\date{October 1, 2026}

\begin{document}
\maketitle
\begin{abstract}
Scuba divers are taught to control their depth to avoid rapid ascents and descents, which could result in serious injuries such as gas embolisms and barotrauma. 
However, many underwater tasks necessitate lateral control, maintaining distance between subsea structures such as coral reefs, submerged drilling instrumentation, or unexploded ordnance. 
In this work, we discuss a first-of-its-kind wearable robotic solution providing thruster-actuated directional guidance to a diver, as distinct from prior propulsive-assistance exoskeletons.
We introduce \textit{Robotic Assisted Diver Movement in Confined Spaces} (RADMCS), a wearable robot that assists divers in maintaining a fixed distance from subsea structures by leveraging perception techniques in monocular depth estimation and force-feedback from submersible thrusters \dchange to provide haptic feedback. \stopchange 
Its small and compact form factor creates a foundational platform that could be expanded to include more sophisticated control and navigation behaviors. 
We present results from Institutional Review Board (IRB) in-water studies with eight human scuba diver participants on threshold sensitivity tests in both a closed-water swimming facility and ocean environments; distance-maintaining experiments in a closed-water facility; and form, fit, and function testing in the ocean. 
We demonstrate that relatively low thrust values ($\approx10$\% of maximum) allow robotic direction of a human's movement using the physical sensation of the robot's guidance. 
\end{abstract}

%%%%%%%%%%%%%%%%%%%%%%%%%%%%%%%%%%%%%%%%%%%%%%%%%%%%%%%%%%%%%%%%%%%%%%%%%%%%%%%%
\section{Introduction}
% Had a reference to amEnde_2001_underwater_cave_mapping
% christian1974swimmer_eod_standoffs
% ,abdullah2024_Cave_Seg
Divers working in confined spaces, such as underwater cave systems~\citep{iliffe2001_cave_diving}, pipelines~\citep{beckwith2012_diving_impact_on_petroleum}, coral reef canyons~\citep{lin2021_touching_coral_reef}, or around unexploded ordnance~\citep{oncticua2024_eod_divers}, must often maintain a safe operating distance from the surrounding walls or objects. 
In the case of cave diving, for example, avoiding cave walls helps divers maintain optimal visibility conditions by avoiding collisions with the surrounding enclosure. 
This minimizes stirring up deposited sediment~\citep{exley1986_basic_cave_diving}, which is a leading cause of diver disorientation-induced fatalities in cave diving accidents~\citep{byrd1997_underwater_cave_fatalities}. 
Maintaining safe standoff distances for divers during coral reef explorations is also important, since ocean currents, other divers, or marine obstacles can inadvertently cause divers to touch coral~\citep{lin2021_touching_coral_reef}, which devastates coral health.
% , touching is to be avoided by keeping safe standoff distances.

\begin{figure}[htb]
\vspace{2mm}
\centering
    \includegraphics[width=1.0\columnwidth]{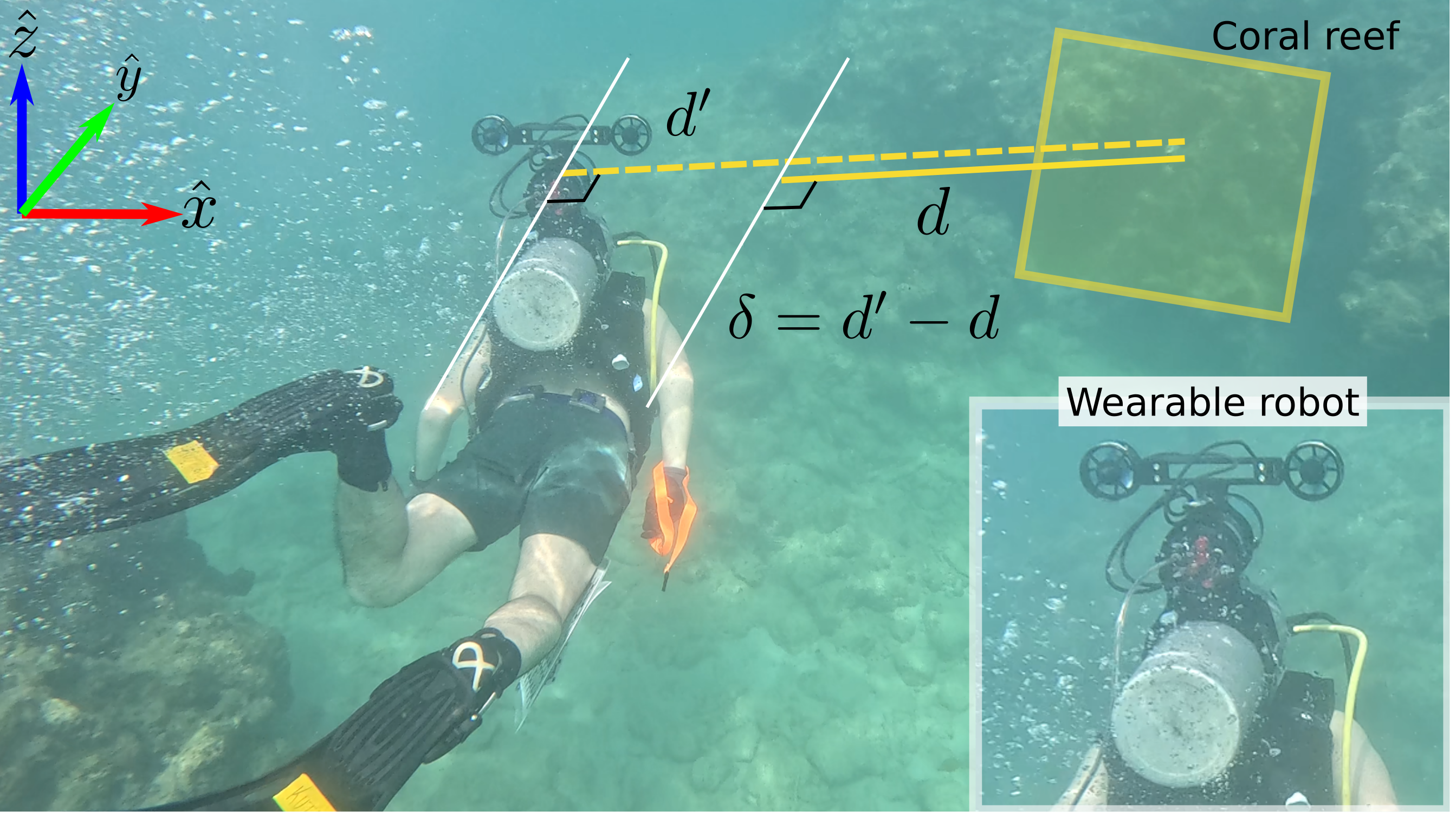}
    \caption{The Robotic Assisted Diver Movement in Confined Spaces (RADMCS) wearable robot during a field experiment. The diver wears RADMCS on their back, enabling \dchange haptic \stopchange feedback from the robot to the human through a series of thruster movements. If the difference $\delta $ between the setpoint distance $d$ and the actual distance of the diver to the coral reef wall $d'$ is greater than zero, the robot indicates that the diver should yaw toward the wall, otherwise yaw away from the wall. The diver responds to the feedback, swimming in the direction indicated by the robot's thrusters, which creates a pathway to complex navigation and maneuvering for safe navigation along coral walls or confined spaces.}
    \label{fig:radmcs_cover_image}
    % \vspace{-5mm}
\end{figure}

In this paper, we introduce RADMCS (\textit{R}obotic \textit{A}ssisted \textit{D}iver \textit{M}ovement in \textit{C}onfined \textit{S}paces) with the goal of addressing these distance-maintaining challenges. 
RADMCS is an easy-to-deploy \textit{wearable} robotic solution that assists in maintaining a preset standoff distance from encompassing walls or objects by providing force-feedback to the diver when they get too close to such objects. 
The force-feedback mechanism is driven by monocular depth estimation techniques that estimate the distance between the robot's cameras and the surrounding structure, and the robot uses its thrusters in response to this perceived distance. 
Fig.~\ref{fig:radmcs_cover_image} \dchange shows the haptic feedback mechanism working in response to the distance between the surrounding reef and the robot's cameras. 
Because the robot is affixed to the diver, the actuation, in turn, perturbs the diver's movements. RADMCS integrates on the diver's existing equipment by strapping onto the scuba diving air tank. 
This eliminates the need for complex exoskeleton style suits or augmented swimming solutions that require additional training or instrumentation. 
It is self-contained and does not interfere with the diver's natural body movements beyond the perturbative feedback provided to the diver.\stopchange

% TODO: Junaed commented this out as it is a duplicate of the text in the above paragraph.
% Because the robot is affixed to the diver, the actuation, in turn, perturbs the diver's movements. 
% RADMCS integrates on the diver's existing equipment by strapping onto the scuba diving air tank. 
% This eliminates the need for complex exoskeleton style suits or augmented swimming solutions that require additional training or instrumentation. 
% It is self-contained and does not interfere with the diver's natural body movements beyond the perturbative feedback provided to the diver.
Our contributions are the
\begin{itemize}
    \item introduction of a novel, wearable underwater robotic system for movement assistance;
    \item presentation of systematic testing of threshold sensitivity (TS) for haptic feedback with human participants in an underwater study in ocean and closed-water environments;
    \item exploration of closed-water experiments of simulated wall-following to gauge the robot's effectiveness at directing human participants;
    \item demonstration of form, fit, and function of the current design in open-water field experiments for TS and maneuverability near coral reefs.
\end{itemize}
\vspace{-2mm}
\section{Literature Review}
This work draws from both wearable guidance in haptics, which is the study of navigation using physical perturbation, and physical underwater human-robot interaction.

% \subsection{Wearable Guidance and Assistive Technology}
\textbf{Wearable Guidance and Assistive Technology}.
Global positioning systems promise precise localization in areas where signals are present. Visual display of GPS navigation information is often inconvenient in instances where people must both navigate and maintain situational awareness of their environment, for example military personnel operating in war zones, visually impaired individuals navigating a busy city center, or individuals driving in unknown locations. An alternative to visual navigation displays are haptic feedback interfaces for wearable guidance.

Wearable guidance refers to a device affixed to the human body that provides some stimuli to assist the human in movement or navigation~\citep{erp2005_waypoint_haptics}. Often the type of stimulus used is vibrotactile~\citep{tsukada2004activebelt_wearable_guidance}, where small motors or actuators can vibrate, effectively encoding messages about direction depending on body mounting positioning. Belt-style wearables arrange tactile actuators in an array, vibrating along specific positions to indicate direction~\citep{tsukada2004activebelt_wearable_guidance}. Alternatively,  in \citet{pielot2010pocketnavigator_haptics} the authors encode heading and distance in duration and rhythm of cell phone vibrations. Others have explored small handheld devices that provide auditory and vibrational feedback in response to receiver-transponder communication~\citep{loomis2005personal_guidance,marston2006evaluation_spatial}. 
% However, the underwater domain does not necessarily benefit from auditory feedback mechanisms, since bubbles from exhalations through scuba diving regulators can cause issues with auditory methods. 
% \vspace{-3mm}

% \input{figs_tex/drysuits}

Immersion protection and thermal layering of undergarments causes issues with vibrational feedback mechanisms, since waterproofing electrical components at the garment level to ensure user safety is critical and challenging. Movement assistance provided by wearable exoskeletons~\citep{golgouneh2021design_soft_exo_skeleton}, even under scuba diving drysuits, could be problematic since the power system and electrical components might fail if the suit floods. Although shape memory alloys (SMAs) and shape metal polymers (SMPs) have been shown to exhibit corrosion resistance and can be submersed for underwater applications~\citep{ma15082841,YANG2024117005}, to our knowledge, there have been no successful deployments of wearable SMAs or SMPs for submersible exoskeletons or rehabilitative robotics.

Some other locations discussed include limb mounting point wearables~\citep{prabhu2020vibrosleeve_sleeve_mounted}, which assist in movement rehabilitation by providing muscular cues about object positioning and distance from the hand, for instance. Feedback is not limited to a single haptic modality. Some employ multiple, potentially interfering, haptic feedback sensors such as vibrotactile combined with compression for limb movement assistance~\citep{subash2025haptic_haptics_vibro_compression}.

% \subsection{Exoskeletons Underwater}
\textbf{Exoskeletons Underwater}. 
Most of the work in physical underwater human-robot interaction has focused on exoskeleton style designs that tightly couple to the diver's body or completely replace scuba diving equipment. These can be summarized as either ``rigid," in the sense that they use metal alloys for forming the external structure; or ``soft," that use cabling or flexible materials for the exoskeleton~\citep{Li2022UnderwaterSurvey}.

% \subsubsection{Rigid Exoskeletons}
\textit{Rigid Exoskeletons}. 
In \citet{Neuhaus2004SwimmingExo_rigid} the authors introduce a conceptual rigid exoskeleton that assists in swimming for reducing metabolic energy consumption. Their suit uses a series of rotary actuators affixed to the diver's legs to provide kicking assistance. They do not discuss the specific implementation of certain life support systems, such as breathing gas integration into the suit. The authors of \citet{zeng2016control} introduce a rigid, back-mounted swimming assistance robot that leverages hydraulic power to move the diver along. The back-mounted unit is sealed to prevent water intrusion. While they achieve good error with planned knee and hip trajectories, the unit occupies the volume of space where the diver's tank would normally go, making it an unsuitable choice for movement assistance. 

Other rigid exoskeletons include the atmospheric diving suit (ADS) and Deep Sea Expeditionary No Decompression (DSEND) suit~\citep{NuytcoExosuitSpec,dsend}, which isolate the diver from the ambient water and pressure. These operate at depths of $300$--$600$~m, without decompression obligations, since the suit maintains the diver at atmospheric pressure. These suits are not robotic, in that they do not have the ability to take action based on environmental observations alone. They are exposure protection suits that allow divers to achieve significant water depths, while maintaining human control via external actuation interfaces. \dchange These suits also cost significantly more than RADMCS, typically on the order of USD $100$K. By comparison, RADMCS costs USD $1$K assembled \cite{ads_cost}.\stopchange

% \subsubsection{Soft Exoskeletons}
\textit{Soft Exoskeletons}. 
 The authors of \citet{wang2020underwater_soft_exoskeleton} introduce a soft exoskeleton that uses a cable and pulley system for transferring rotational power to the swimmer's ankle joints. However, the purpose of the exoskeleton is limited to rehabilitative swimming, since it occupies much of the diver's lower limbs and upper torso. It is not clear how breathing gas would be provided to a scuba diver wearing their exoskeleton. The authors of \citet{wang2025system_soft_exoskeleton} introduce a soft flexible exoskeleton that accommodates a diver's regular buoyancy control vest and scuba tank, making it a more realistic alternative for actual scuba diving operations. They too use a cable system with two hip mounted actuator winches for cable movement. Their suit is meant for increasing swimming efficiency. Additional cabling and pulleys, while functional, often add additional pinch points or potential entanglement hazards for divers working in hazardous conditions. In contrast, the RADMCS robot integrates on a simple scuba breathing gas cylinder, which eliminates the issues with cable and pulley systems.
\section{System Design}
\label{ch:radmcs_system_design}
\begin{figure*}[thbp]
    \includegraphics[width=1.0\textwidth]{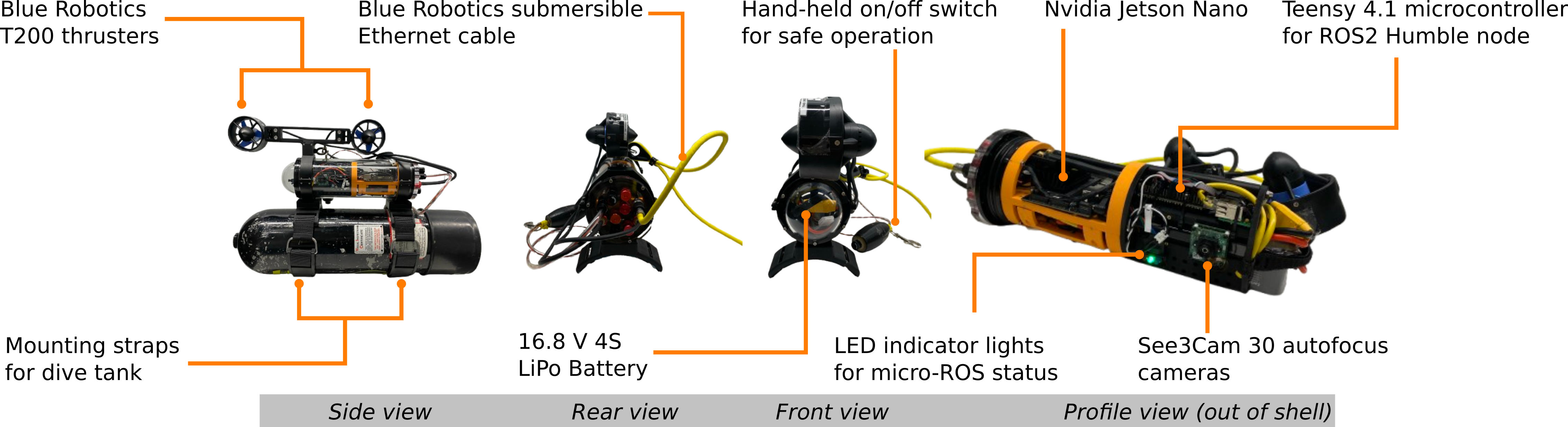}
    \caption{Assembled RADMCS robotic system features with descriptions of parts.}
    \label{fig:ch6_radmcs_with_descriptions}
    % \vspace{-3mm}
\end{figure*}

Unlike exoskeletons that attach to the diver's body directly, the proposed design of the RADMCS robot enables a less invasive integration with existing scuba diving technology. Namely, the RADMCS robot mounts onto the diver's scuba tank using strapping mechanisms already employed by open-water scuba divers. The design comprises two primary components: the external assembly, which has the mechanism for strapping to the diver; and the internal assembly, which houses the electronics subassembly and sensing technology. 
Fig.~\ref{fig:ch6_radmcs_with_descriptions} shows these components on the assembled system.

\begin{figure}[thbp]
\centering
    \includegraphics[width=0.75\columnwidth]{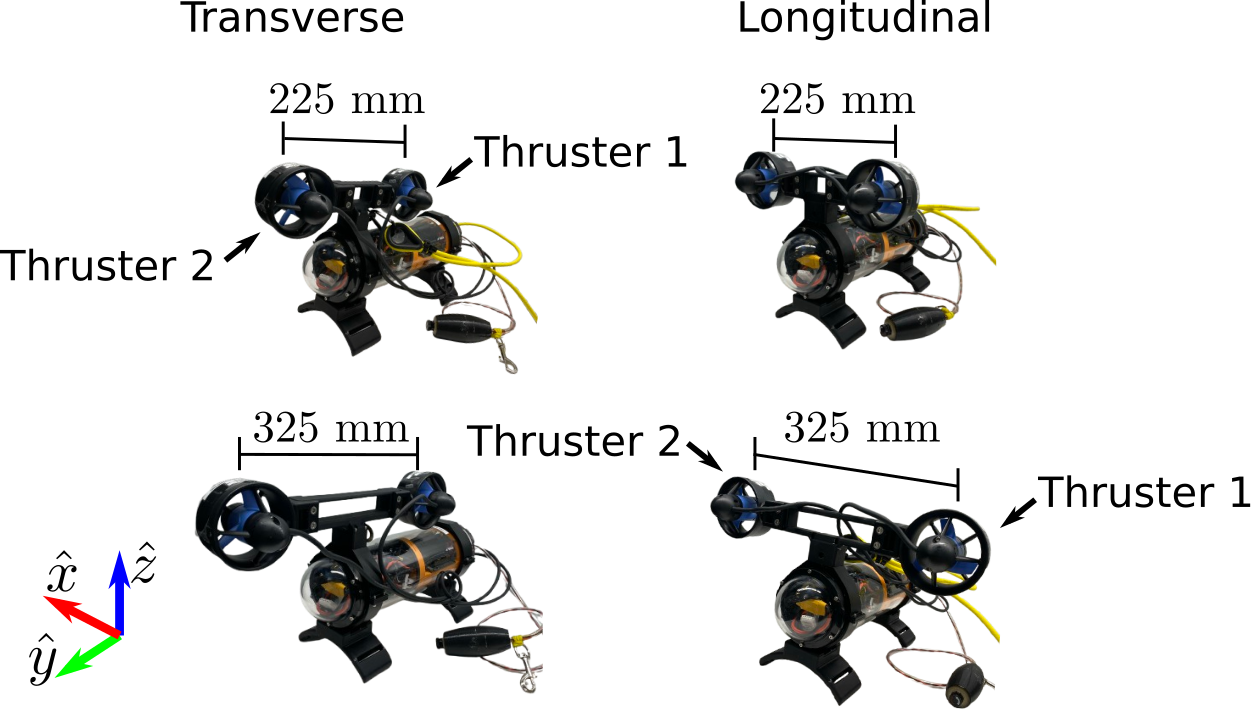}
    \caption{Thruster configurations. The left image column shows the transverse thruster configuration that directs force perpendicular to the robot's body, in the $\hat{x}$-direction. The right image column shows the longitudinal thruster configuration that directs force along the length of the robot's body, in the $\hat{y}$-direction. The thrusters are labeled and referred to as shown in the figure for all future analysis. }
    \label{fig:ch6_radmcs_thruster}
    % \vspace{-2mm}
\end{figure}

% \subsection{External Assembly}
\label{ch6:radmcs_external_assembly}
\textbf{External Assembly}. 
The robot's external assembly consists of a single cylindrical acrylic shell, $300$~mm in length and $102$~mm in diameter. \dchange The shell is depth rated to $100$~m by the manufacturer \citep{blueRobotics_enclosure}, making RADMCS suitable for operational testing within in-shore commercial and practical recreational diving limits \citep{adci_standards}. \change
\dchange The robot is affixed to the diver's scuba tank using 3D printed parts that match the curvature of a common ``aluminum 80" scuba diving cylinder, which has a diameter of $185$~mm. \stopchange 
This design could easily be adapted for other scuba diving tank dimensions by modifying the 3D printed tank attachment interface.
The robot is actuated using two thrusters (Blue Robotics T200 ~\citep{blueRobotics_thrusters}). 
Specific thruster configurations can be achieved by interchanging the thruster interface module. 
This uses an additively-manufactured part to bolt onto the robot's clamp. 
A specific design consideration is the extent to which the thruster configuration affects maneuverability for the diver and sensitivity to perturbations. 
See Fig.~\ref{fig:ch6_radmcs_thruster} for specific thruster configurations and naming conventions. 
In the \textbf{longitudinal} configuration, thrust acts along the robot’s $\hat{y}$-axis, and in the \textbf{transverse} configuration, thrust acts along the $\hat{x}$-axis. We explore these design choices during closed-water and open-water field experiments.

Data transfer and control links, \eg for testing and experiments, is established through a neutrally-buoyant gigabit Ethernet tether (Blue Robotics Fathom tether coupled with  an 8-pin~\citep{bluetrail} bulkhead connector).
However, the device does not require a tether for regular operations.
% An external, hand-held switch (shown as ``on/off switch'' in Fig.~\ref{fig:ch6_radmcs_with_descriptions}) enables human-controllability while the robot is affixed to the diver's back. This is a safety feature to ensure that if the robot acts in an unexpected manner, the user can manually power the system off during operations with or without the tether.

% \subsection{Internal Assembly}
\label{ch6:radmcs_internal_assembly}
\textbf{Internal Assembly}. 
The internal assembly consists of two interlinking 3D printed cylindrical sections.
The aft-section mounts to a flange assembly and provides sufficient spacing for pass-through (WetLink) and potted penetrators. 
The aft-section bolts to the fore-section to create a robust assembly. 
The fore-section provides slotting for the robot's battery and cameras. 
A single $14.8$~V $6500$~mAh $4$S Lithium Polymer (LiPo) battery supplies the computational module, the thrusters, the microcontroller, and the cameras. \dchange The worst case scenario for power consumption (bounded by the thruster power requirements) is $20$ minutes of runtime with a single battery, given manufacturer-provided current draw at maximum thrust for two thrusters ($\approx40$~A). We achieve significantly better duration by using intermittent haptic actuation, often exceeding $4$~hours.\stopchange

Currently, two autofocus RGB cameras provide vision for the robotic system. % (See3Cam\_30 by E-Con Systems~\citep{econcameras})
The cameras are mounted on the sides of the robot (left/right), facing the direction of the surrounding walls or structures. A Teensy $4.1$ microcontroller provides the output PWM signal to the thrusters. \dchange A \textit{micro-ros} client and agent running \stopchange
ROS 2 Humble \citep{ros2} enable communication between software components. 
% A small LED indicator light communicates visual feedback of the connectivity state between the \textit{micro-ros} client and \textit{micro-ros-agent}.

% \subsection{Haptic Feedback Mechanism}
\label{sec:radmcs_haptic_feedback}
\textbf{Haptic Feedback Mechanism}. 
Robotic feedback is provided by controlling thrust actuation. 
Each thruster is controlled by a pulse-width modulation (PWM) signal from $1500~\mu$s at neutral to $1900~\mu$s in the full forward thrust or $1100~\mu$s at full reverse thrust. 
Feedback is delivered to the human via the physical sensation of the robot moving the diver in the intended direction. 
Let $T1$ and $T2$ denote the PWM values for thruster 1 and thruster 2, respectively.
The robot indicates the diver should turn left, or yaw counterclockwise (CCW) about the $\hat{z}$-direction, using $T1\in[1100,1500]~\mu$s reverse and $T2\in[1500,1900]~\mu$s forward. 
Conversely, the robot indicates the diver should turn right, or yaw clockwise (CW) about the $\hat{z}$-direction, using $T1\in[1500,1900]~\mu$s forward and $T2\in[1100,1500]~\mu$s reverse.
\begin{table*}[ht]
\caption{Directional movement mapping of the haptic feedback from robot to human. In both the longitudinal and transverse thrust configurations the movement instruction is the same.}
\setlength{\tabcolsep}{8pt} % Adjusts space between columns
\renewcommand{\arraystretch}{1.2} % Increases row height
\centering
\begin{tabular}{l | l}
\toprule
\textbf{Movement} & \textbf{Longitudinal/Transverse}\\
\specialrule{0.8pt}{0pt}{0pt} % Replaces \toprule
Turn left (yaw CCW about $\hat{z}$) & $T1$ reverse and $T2$ forward \\
Turn right (yaw CW about $\hat{z}$) & $T1$ forward and $T2$ reverse \\
\specialrule{0.8pt}{0pt}{0.8pt} % Replaces \toprule
\end{tabular}
% \vspace{-2mm}
\label{table:radmcs_movement_mapping}
\end{table*}
% \vspace{-2mm}
\section{Visual Control Methodology}
\label{sec:radmcs_metholodology}
We employ a linear control algorithm with saturation to limit our thruster outputs after a certain distance. Alg.~\ref{alg:radmcs_control} outlines the pseudocode for the control algorithm. The goal of the algorithm is to produce thruster PWM values in microseconds for $T1$ and $T2$. Let $I$ be an image of arbitrary dimensions. Our control algorithm computes the distance $d^i$ between the camera and the target using the \textproc{FindDistance} function. \dchange A target can be any specific object or feature from which the algorithm can extract distance. For example,\stopchange \citet{abdullah2024_Cave_Seg} uses deep learning-based methods for extracting features from cave walls. For our single-sided distance-maintaining experiment in Section~\nameref{ch6:radmcs_single_sided_test}, we used AprilGrids from the \textit{Kalibr} specification~\citep{oth2013rolling}, arranged in a $4\times6$ grid, with tag size $50$~mm, and tag spacing $0.30$, encoded with the $t36h11$ tag dictionary. This ensured semi-stable distance estimates. We designed the RADMCS robot to work with any method that computes distance between the robot's cameras and the surrounding enclosure. To decouple the efficacy of the proposed haptic mechanism from monocular distance estimation, we only consider existing, classical methods for distance estimation. Stereo-based, deep learning methods, or other feature-based methods could also be employed. This is the subject of future work.
% \vspace{-1mm}

In this case, to extract distance estimates, we leverage classical Perspective-n-Point estimation (PnP) to recover camera pose from known 3D positions specified by the AprilGrid and known 2D image locations~\citep{bradski2008learning}. However, even with stable distance estimates, underwater distortion, lighting variations, and autofocus features created instabilities in our distance estimates $d'$, which we empirically observed to exceed $+4$~m at times. To partially offset these challenges, we implemented an exponential moving average (EMA) filter to smooth our distance estimates and avoid large unphysical jumps in distance estimates (line~\ref{alg1:line_ema} in Alg.~\ref{alg:radmcs_control}). Additionally, we implemented an empirically determined maximum likely distance heuristic $D_\text{MAX}$ that prevented filter updates if the difference between our old estimate $d'$ and the current distance $d^i$ exceed this threshold. The EMA was configured with $\alpha=0.5$, providing modest filter response.

Additionally, we created a control deadzone $\delta_\text{deadzone}$ in which the robot would not provide haptic feedback, because the diver was effectively traversing without the need for movement assistance. We compute the control error $\delta$ as the difference between the setpoint $d$ and the current filtered distance estimate $d'$, \ie $\delta=d'-d$.
% \vspace{-6mm}
% \begin{equation}
%     \delta=d-d'_\text{filtered}.
% \end{equation}
\begin{algorithm}[tbhp]
\caption{Linear proportional control with saturation.}\label{alg:radmcs_control}
\begin{algorithmic}[1]
\Function{$[T1,T2]~=$~Control}{$I$}
    \State{$d'$= null}\Comment{Filtered distance initialization}
    \State $d_i\gets$ \Call{FindDistance}{$I$}\Comment{Distance estimation}
    \If{$d' =$~null}~\State{$d'=d_i$}
    \ElsIf{$\lvert d'-d_i\rvert < D_{\text{MAX}}$}
         \State{$d'=\alpha \cdot d_i+(1-\alpha) \cdot d'$}\label{alg1:line_ema}\Comment{EMA update}
    \EndIf
    \State{$\delta=d'-d$}\Comment{Setpoint error computation}
    \If{$\lvert\delta\rvert\leq\delta_{\text{deadzone}}$} \Return $[1500,1500]$\Comment{Neutral}
    \Else
        \State{$\delta\gets$\Call{Clip}{$\delta,D_{\text{range}}$}}\Comment{Saturation}
        \State{$T1=$~\Call{ComputePWM}{$\delta,T1_{\text{MAX}},T1_{\text{MIN}}$}}
        \State{$T2=$~\Call{ComputePWM}{$\delta,T2_{\text{MAX}},T2_{\text{MIN}}$}}
        \If{$\delta<0$}
        \Return $[T2,T1]$ \Comment{\textit{Turn CCW}}
        \Else~\Return $[T1,T2]$ \Comment{\textit{Turn CW}}
        \EndIf
    \EndIf
\EndFunction
\Function{$x=$~Clip}{$\delta,D_{\text{range}}$}
    \State{$x=\delta/D_{\text{range}}$}
    \If{$x<-1$}
        \Return $-1$
    \ElsIf{$-1\leq x \leq 1$}~
        \Return $x$
    \Else~
        \Return $1$
    \EndIf
\EndFunction
\Function{$T=$~ComputePWM}{$\delta,T_{\text{MAX}},T_{\text{MIN}}$}
    \State{$c=\frac{T_{\text{MAX}}+T_{\text{MIN}}}{2}; s=\frac{T_{\text{MAX}}-T_{\text{MIN}}}{2}$}
    % \State{$s=\frac{T_{\text{MAX}}-T_{\text{MIN}}}{2}$}
    \State{$T=c+\delta\cdot s$}
    
    \Return $T$
\EndFunction
\end{algorithmic}
%\vspace{-4mm}
\end{algorithm}
We clip this value to a practical control range $D_\text{range}$. This is because the cameras (using any practical \textproc{FindDistance} method) have a limited sensing range. In our case, we determined that the cameras and our PnP algorithm had a useful range of $D_\text{range}=1.5$~m. Our clipping function, saturates the control, so that any distance estimate that exceeds $D_\text{range}$ is set to the maximum allowable thruster value, defined as $T_\text{MAX}$, or the minimum $T_\text{MIN}$. The \textproc{Clip} function is defined as $f(x) = \{-1, \text{if } x < -1; x, \text{if } -1 \leq x \leq 1; 1, \text{if } x > 1\}$. This scales our error by $D_\text{range}$ and saturates at the limits $\delta=\pm D_\text{range}$. Finally, the function \textproc{ComputePWM} ensures the PWM values are limited to the maximum $T_\text{MAX}$ and minimum $T_\text{MIN}$ thruster values. These parameters are derived from the TS tests to gauge the starting PWM $T_\text{MIN}$ that is at a level the user can feel. We can also limit the upper bounds of the thruster force by setting $T_\text{MAX}$. This can be tailored to specific use cases, \eg when extended battery life is required, a lower $T_\text{MAX}$ value requires less power for the same period of operational time.
% \vspace{-1mm}
% \vspace{-2mm}
% \begin{equation}
%     f(x) = 
% \begin{cases}
%     -1, & \text{if } x < -1 \\
%     x,  & \text{if } -1 \leq x \leq 1 \\
%     1, & \text{if } x > 1.
% \end{cases}
% \end{equation}
\section{Human Study Evaluations}
\label{ch6:radmcs_human_study_setup}
% approvals STUDY00027283 and MOD00060722 at the University of Minnesota--Twin Cities, 
Under Institutional Review Board (IRB) approval, we conducted in-water human studies to evaluate the RADMCS system. Eight people with varying levels of scuba diving experience participated in a variety of in-water experiments to gauge TS and single-sided distance-maintaining wall following. \dchange While the number of participants is lower compared to terrestrial haptics studies (see \cite{erp2005_waypoint_haptics} which used $12$ participants), conducting underwater human studies creates a significant barrier to entry compared to its terrestrial counterpart. These challenges include: recruitment of certified and qualified individuals (scuba divers are not a typical demographic); planning for approval and consent processes (to mitigate the inherent safety risk); swimming facility and equipment rental coordination (large overhead to rent time in a swimming facility, acquire dive equipment, or deploy in an ocean); along with acquiring support staff to ensure participant safety (safety divers to monitor participants for injuries or behaviors that could lead to injuries). These factors make this portion of our work significant, representing a step forward for underwater haptics and wearable robotics research. \stopchange Table~\ref{table:participant_table} shows the specific portions of the study completed by each participant. Participants are referred to as P1--P8 for the purposes of our analysis.

% \begin{table*}[ht]
% \caption{Participant list and the specific portions of the study they were involved in. Participants are coded from P1--P8.}
% % \setlength{\tabcolsep}{5pt}
% \renewcommand{\arraystretch}{1.1}
% \newcolumntype{Y}{>{\centering\arraybackslash}X}
% \small
% \begin{tabularx}{\textwidth}{l | Y | Y | Y | Y | Y}
% \toprule
% \textbf{Participant/Event} & 
% \textbf{Threshold sensitivity closed-water $225$~mm longitudinal and transverse} & 
% \textbf{Threshold sensitivity closed-water $325$~mm transverse} & 
% \textbf{Single-sided distance-maintaining} & 
% \textbf{Threshold sensitivity open-water $325$~mm longitudinal and transverse} & 
% \textbf{Open-water form, fit, \& function} \\
% \specialrule{0.8pt}{0pt}{0pt}
% \rowcolors{2}{blue!10}{white}
% P1 & $\checkmark$ & & $\checkmark$ & $\checkmark$ & \\%Rishi
% P2 & $\checkmark$ & & $\checkmark$ & & \\%Sakshi
% P3 & & & & $\checkmark$& \\%David
% P4 & & & $\checkmark$& & \\%Ayaan
% P5 & & $\checkmark$& $\checkmark$& & \\%Merrick
% P6 & & & $\checkmark$& & \\%Jenna
% P7 & & $\checkmark$ & $\checkmark$& & \\%Andrew
% P8 & & & $\checkmark$ & & $\checkmark$ \\%Demetri
% \specialrule{0.8pt}{0pt}{0.8pt}
% \end{tabularx}

% \label{table:participant_table}
% \end{table*}

%% Single column version
\begin{table*}[ht]
\caption{Participant list and the specific portions of the study they were involved in. Participants are coded from P1--P8. Threshold sensitivity testing is abbreviated as (TS).}
\renewcommand{\arraystretch}{1.1}
\newcolumntype{Y}{>{\centering\arraybackslash}X}
\small
\begin{tabularx}{\textwidth}{m{5.40cm} |cccccccc }
\toprule
\textbf{Event/participant} & P1 & P2 & P3 & P4 & P5 & P6 & P7 & P8\\
% \specialrule{0.8pt}{0pt}{0pt}
% \rowcolors{2}{blue!10}{white}
\hline
\textbf{TS closed-water $225$~mm longitudinal/transverse} & $\checkmark$ & $\checkmark$ & & & & & & \\
\hline
\textbf{TS closed-water $325$~mm transverse} & & & & & $\checkmark$ & & $\checkmark$ &  \\
\hline
\textbf{Single-sided distance-maintaining} & $\checkmark$ & $\checkmark$ & & $\checkmark$ & $\checkmark$ & $\checkmark$ & $\checkmark$ & $\checkmark$\\
\hline
\textbf{TS open-water $325$~mm longitudinal/transverse} & $\checkmark$ & & $\checkmark$ & & & & & \\
\hline
\textbf{Open-water form, fit, \& function} & & & & & & & & $\checkmark$ \\
\specialrule{0.8pt}{0pt}{0.8pt}
\end{tabularx}

\label{table:participant_table}
% \vspace{-4mm}
\end{table*}

% Fig.~\ref{fig:radmcs_human_study_plots} shows a breakdown of the expertise and qualification levels of the participants.

We designed a series of incremental in-water tests to evaluate the RADMCS robot, beginning with detection threshold identification and ultimately progressing to submerged navigation tasks. To determine the minimum thrust level at which divers can perceive robotic feedback (\ie threshold sensitivity), we used a standard method of limits procedure \citep{jones2013_haptic_threshold}. We conducted TS  testing in both closed-water and open-water settings, to gauge the difference between environments. We then created a simulated wall-following test to gauge how well the robot provides feedback to the human, \ie correct movement instructions based on distance to the wall.
% Due to logistical constraints, not every individual could participate in every event.
% Rishi
% Ayaan
% Merrick
% Jenna
% Sakshi
% David
% Andrew 

% \subsection{Closed-water Threshold Sensitivity} 
\label{ch6:radmcs_threshold_testing}
\textbf{Closed-water Threshold Sensitivity}. 
We tested the robot using a data transfer tether, which allowed us to sweep pulse-width modulation (PWM) values during ``ascending'' trials, from neutral ($1500~\mu s$) to maximum thrust ($1900~\mu s$); and from maximum to neutral during ``descending.'' We did this for four study participants in the pool in three separate thruster configurations, including the $225$~mm longitudinal and transverse configurations and the $325$~mm transverse configuration. Due to testing facility time restrictions, we did not collect sensitivity data in the longitudinal $325$~mm thruster configuration.   

\begin{figure}[tbp]
\vspace{2mm}
\centering
    \includegraphics[width=1.0\columnwidth]{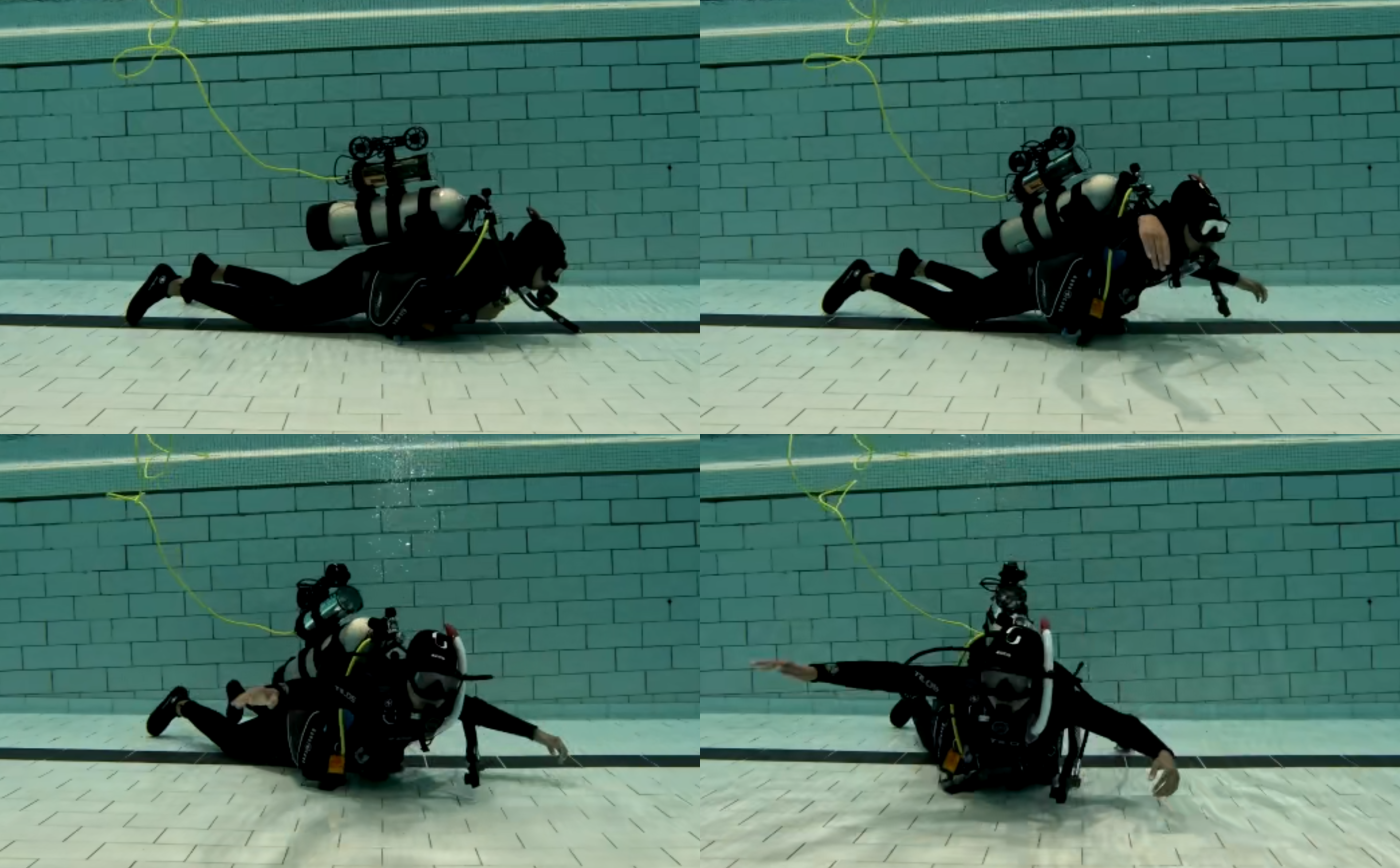}
    \caption{Still frames collected during the closed-water TS test for the $225$~mm transverse thruster configuration. During the ``ascending'' portion of the test, the thruster force-feedback increases in intensity, rotating the diver in the figure.}
    \label{fig:radmcs_threshold_stills}
    % \vspace{-3mm}
\end{figure}

\begin{figure}[htb!]
    \centering  
    
    \subfloat[CW rotation/longitudinal thruster configuration.]{%
        \label{fig:cw_longitudinal}%
        \includegraphics[width=0.60\columnwidth]{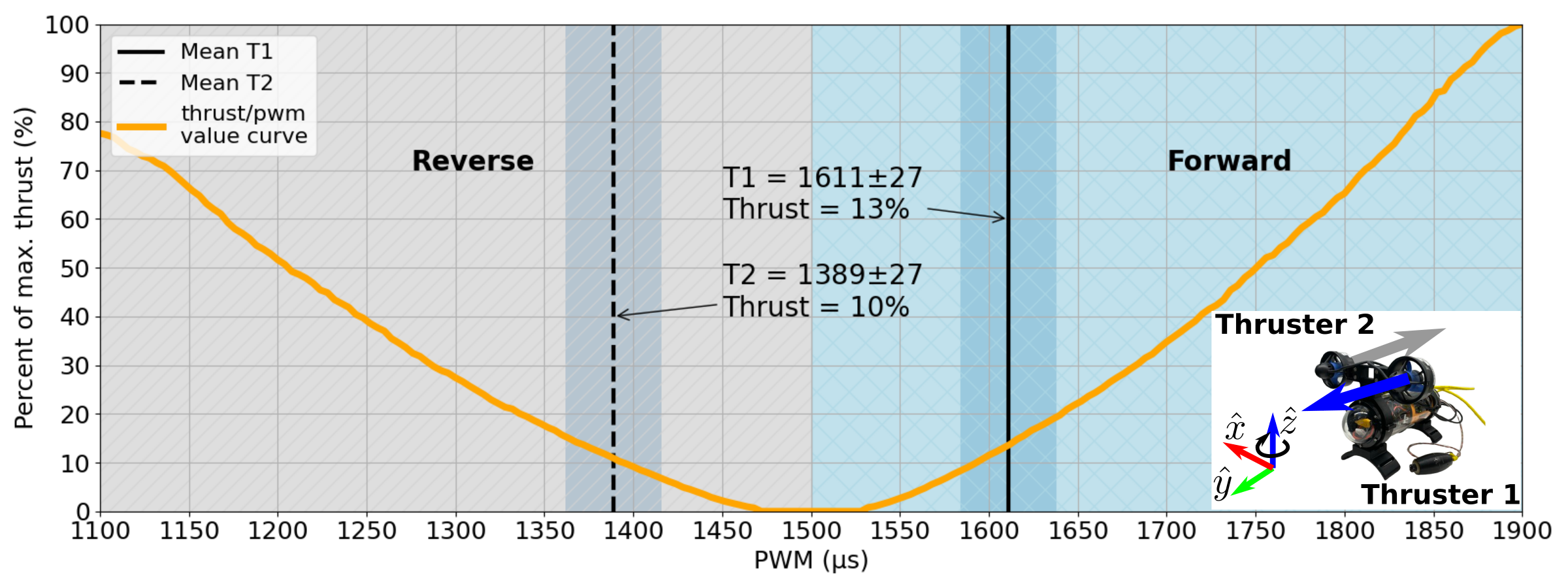}%
    }
    
    \subfloat[CCW rotation/longitudinal thruster configuration.]{%
        \label{fig:ccw_longitudinal}%
        \includegraphics[width=0.60\columnwidth]{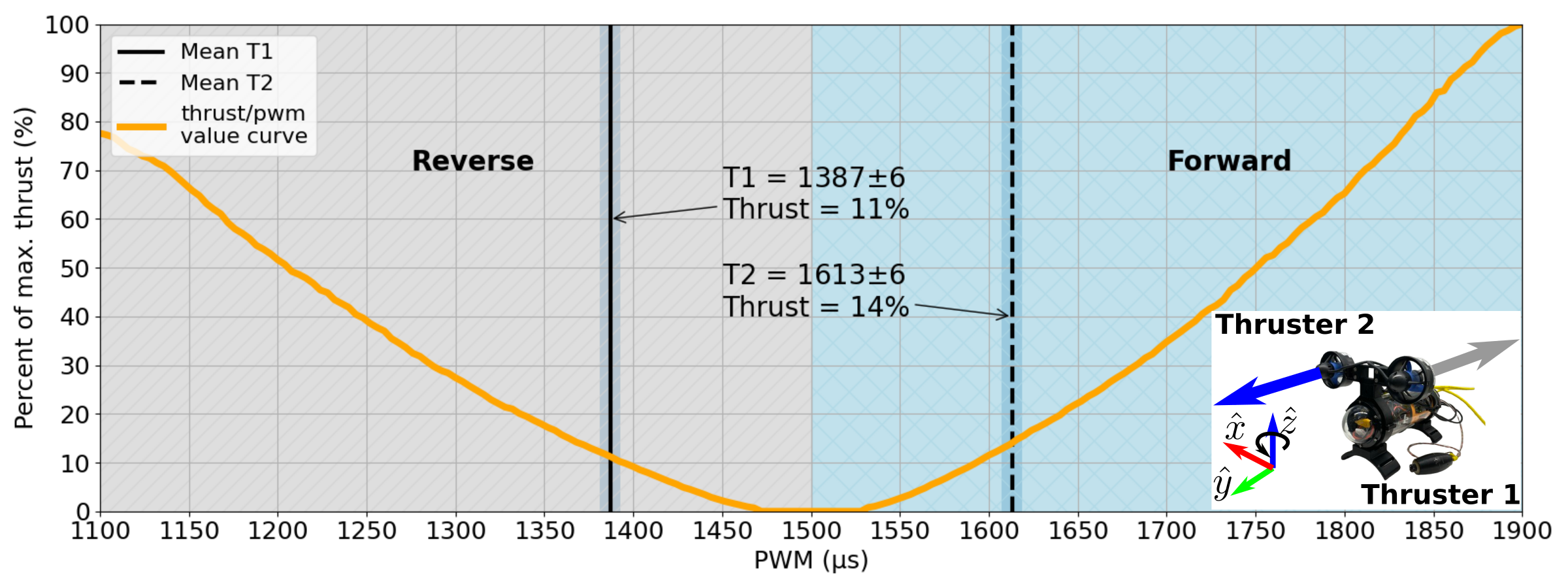}%
    }

    \subfloat[CW rotation/transverse thruster configuration.]{%
        \label{fig:cw_transverse}%
        \includegraphics[width=0.60\columnwidth]{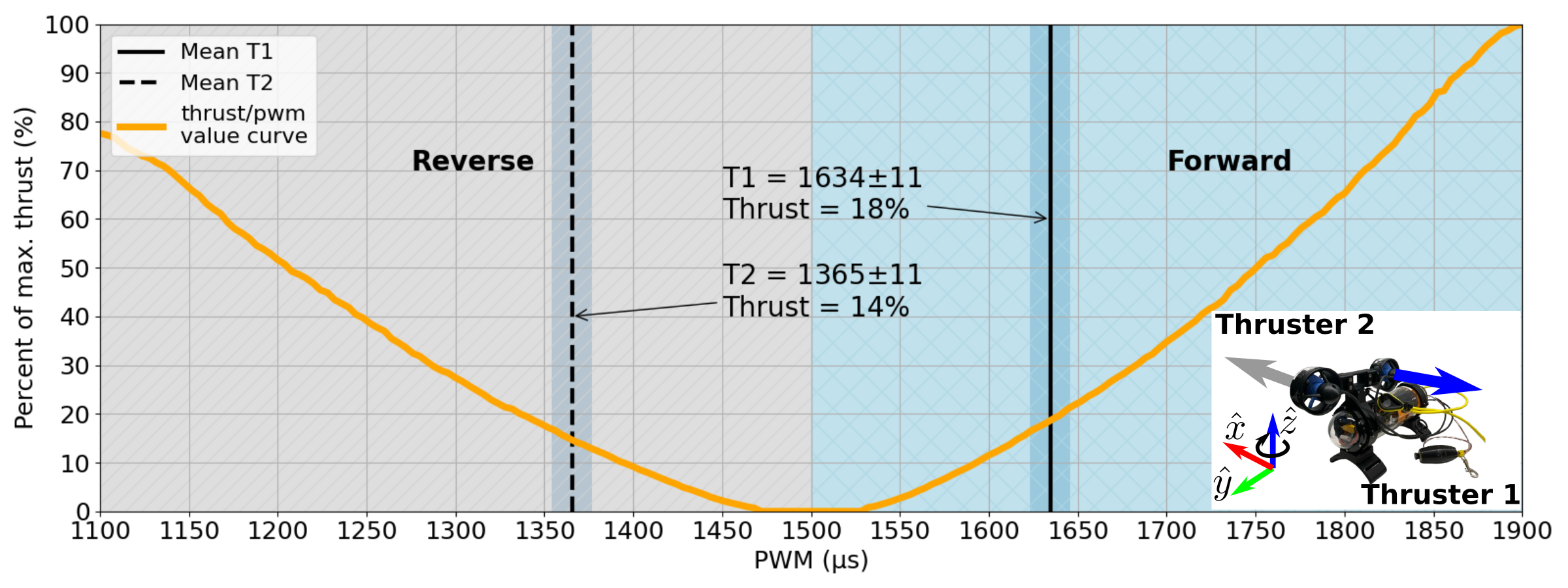}%
    }
    
    \subfloat[CCW rotation/transverse thruster configuration.]{%
        \label{fig:ccw_transverse}%
        \includegraphics[width=0.60\columnwidth]{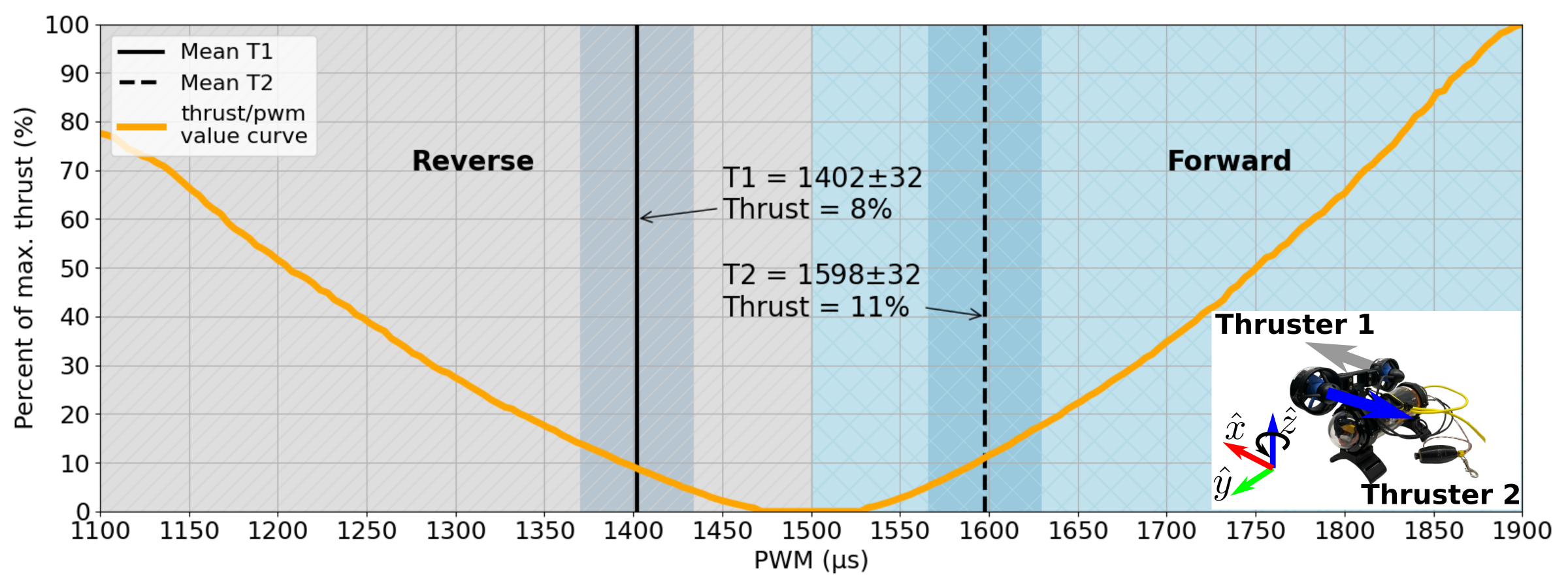}%
    } 
    
    \caption{TS for ($M=2$) participants P1 and P2 for CW and CCW rotations in both the $225$~mm longitudinal and transverse thruster configurations in the closed-water environment. The gold solid curve shows the manufacturer-provided thrust–PWM calibration \citep{blueRobotics_thrusters}. The inset illustrates the thruster orientations and resulting force directions used to generate rotational motion.}
    \label{fig:radmcs_pool_225mm}
    % \vspace{-3mm}
\end{figure}

Each participant wore the RADMCS robot while neutrally buoyant at a depth of $1.8$~m in a controlled pool environment. Participants indicated the moment they first perceived force-feedback. Fig.~\ref{fig:radmcs_threshold_stills} shows an example of the diver participating in the ascending portion of the test. The robot's force-feedback increases, rotating the diver in place. Fig.~\ref{fig:radmcs_pool_225mm} shows the results from the detection threshold test for $225$~mm longitudinal and transverse thruster configuration for the clockwise (CW) and counterclockwise (CCW) rotation directions. Each plot shows the gold thrust vs. PWM curve provided by Blue Robotics~\citep{blueRobotics_thrusters} for the T200 thruster powered at $16.8$~V, spanning from $1100~\mu$s at maximum reverse thrust to $1900~\mu$s at maximum forward thrust. We plot the reverse region with gray dashed highlighting, and we plot the forward region with blue crossed highlighting. For $M$ participants total, let $j$ denote the participant, where $j\in\{1,\dots,M\}$, and let $T1^j$ and $T2^j$ denote the values of thruster 1 and thruster 2 for participant $j$, respectively. Then the solid black line in the plot shows the mean across participants for each thruster $T$. For example, the mean across participants for $T1$ is given by $\langle T1\rangle  = \frac{1}{2M}\sum_{j}(T1^j_\text{ascending}+T1^j_\text{descending})$. The same equation applies to the $T2$ mean. Finally, we also show a shaded region around the mean values that indicates the spread up to one standard deviation to quantify the uncertainty in each measurement. We also report the nearest whole thrust value for each mean, to assess the relative thrust percentage required in each case.
% \begin{equation}
%     \langle T1\rangle  = \frac{1}{2M}\sum_{j}(T1^j_\text{ascending}+T1^j_\text{descending}).
% \end{equation}

% Likewise Fig.~\ref{fig:radmcs_pool_325mm} shows the results from the detection threshold test for $325$~mm longitudinal and transverse thruster configuration for the clockwise (CW) and counterclockwise (CCW) rotation directions. 

Across four participants and all three configurations, the averaged ascending and descending detection thresholds correspond to approximately $10\%$ of the maximum available thrust, substantially lower than anticipated. Additionally, there is no substantial difference between the length of the torque arm (\ie $225$~mm versus $325$~mm) for either configuration. While not conclusive, this finding establishes a clear baseline for continued efforts that leverage any configuration with low thrust levels to start, since even modest actuation produces perceptible directional cues, suggesting that RADMCS can leverage very small perturbations to communicate navigational information without imposing excessive forces on the diver.

% \input{figs_tex/threshold_sensitivity_200mm_trans_pool}

% \input{figs_tex/pool_trial_two}

\begin{figure}[htbp]
\centering
    \includegraphics[width=1.0\columnwidth]{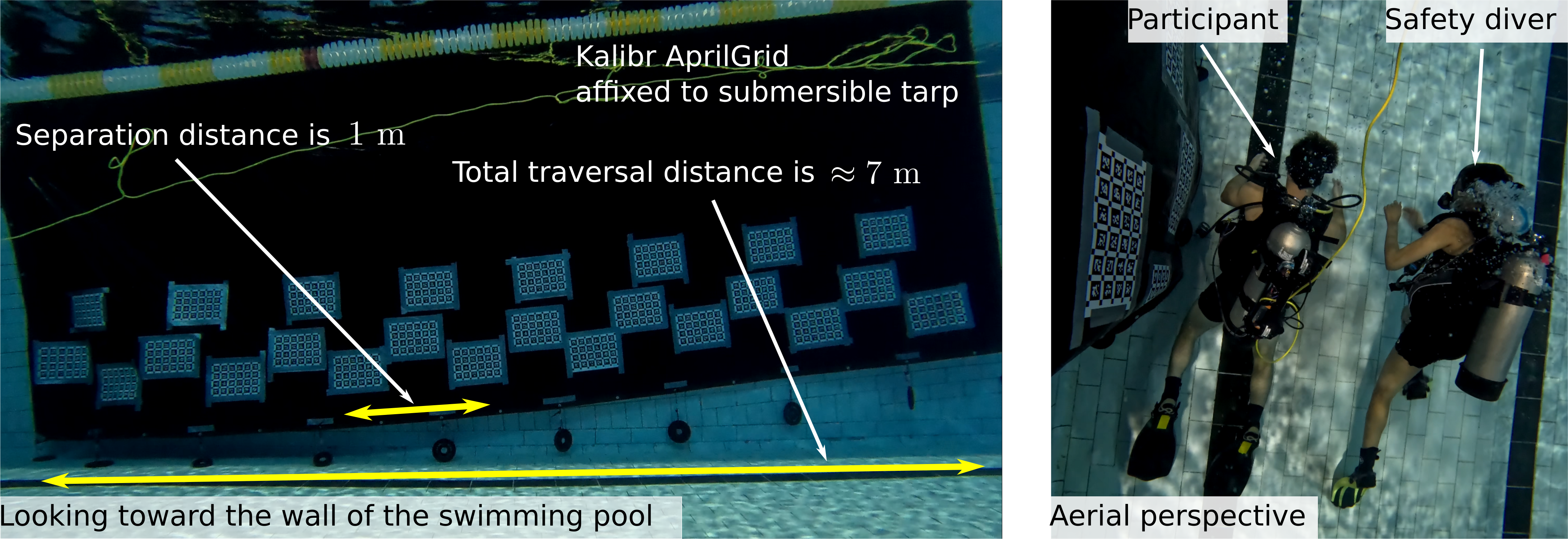}
    \caption{Simulated distance-maintaining experimental setup in the closed-water swimming facility. A tarp with \textit{Kalibr} AprilGrids was affixed to the side of the pool. The currents produced from circulation jets created an imperfect surface leading to curvature and distortions in the images. These curvatures along with facets of the camera (including continual autofocus) had downstream effects on the quality of the control algorithm, since the feedback signal is the distance between the robot's cameras and the wall.}
    \label{fig:ch6_radmcs_pool_trial_setup}
    % \vspace{-3mm}
\end{figure}

% \subsection{Closed-water Single-Sided Distance-maintaining}
\label{ch6:radmcs_single_sided_test}
\textbf{Closed-water Single-Sided Distance-maintaining}. 
To assess if the RADMCS robot can help a diver maintain a safe standoff distance from a subsea structure, we conducted two separate trials in a closed-water swimming facility over two days. We created a semi-controlled experimental setup leveraging the side of the swimming facility. 
% Fig.~\ref{fig:ch6_radmcs_pool_trial_two} shows the schematic and control parameters definitions based on the formulation in Alg.~\ref{alg:radmcs_control}. 
We deployed a submersible tarp with \textit{Kalibr} AprilGrids affixed in a structured pattern, separated by $1$~m, spanning a length of $7$~m. Fig.~\ref{fig:ch6_radmcs_pool_trial_setup} shows the tarp in the left hand image, looking toward the side of the swimming pool wall. Unfortunately, due to depth changes over the length of the tarp, the targets did not remain level with respect to the bottom of the pool, creating the incline visible in the figure. Currents from the circulation jets caused significant warping of the tarp, which caused sporadic distance estimations. 
% \dchange The robot was configured to have only one camera active, on the left or port side of the robot. 
% This is because preliminary laboratory testing revealed that the primary computational module, which is an Nvidia Jetson Nano~\citep{nvidiacomp} GPU hardware, could not run both camera drivers and perform real-time distance computations without significant bandwidth issues. \stopchange

We also employed an aerial trackline camera centered on the black lane marker to capture an aerial perspective. The right-hand image in Fig.~\ref{fig:ch6_radmcs_pool_trial_setup} shows an example aerial perspective. Each participant diver was asked to wear a ``blackout'' mask, completely occluding their own vision. These masks are used by cave divers to train for zero visibility conditions~\citep{iliffe2001_cave_diving}. Each participant diver was instructed to orient themselves along the black line, which is a meter from the pool wall. They were then instructed to put on their blackout mask. The safety diver accompanied the participant diver the entire length of the tarp, intervening when necessary to ensure the participant diver did not swim more than a meter perpendicularly to the direction of travel.

\begin{figure}[htbp]
\centering
    \includegraphics[width=1.0\linewidth]{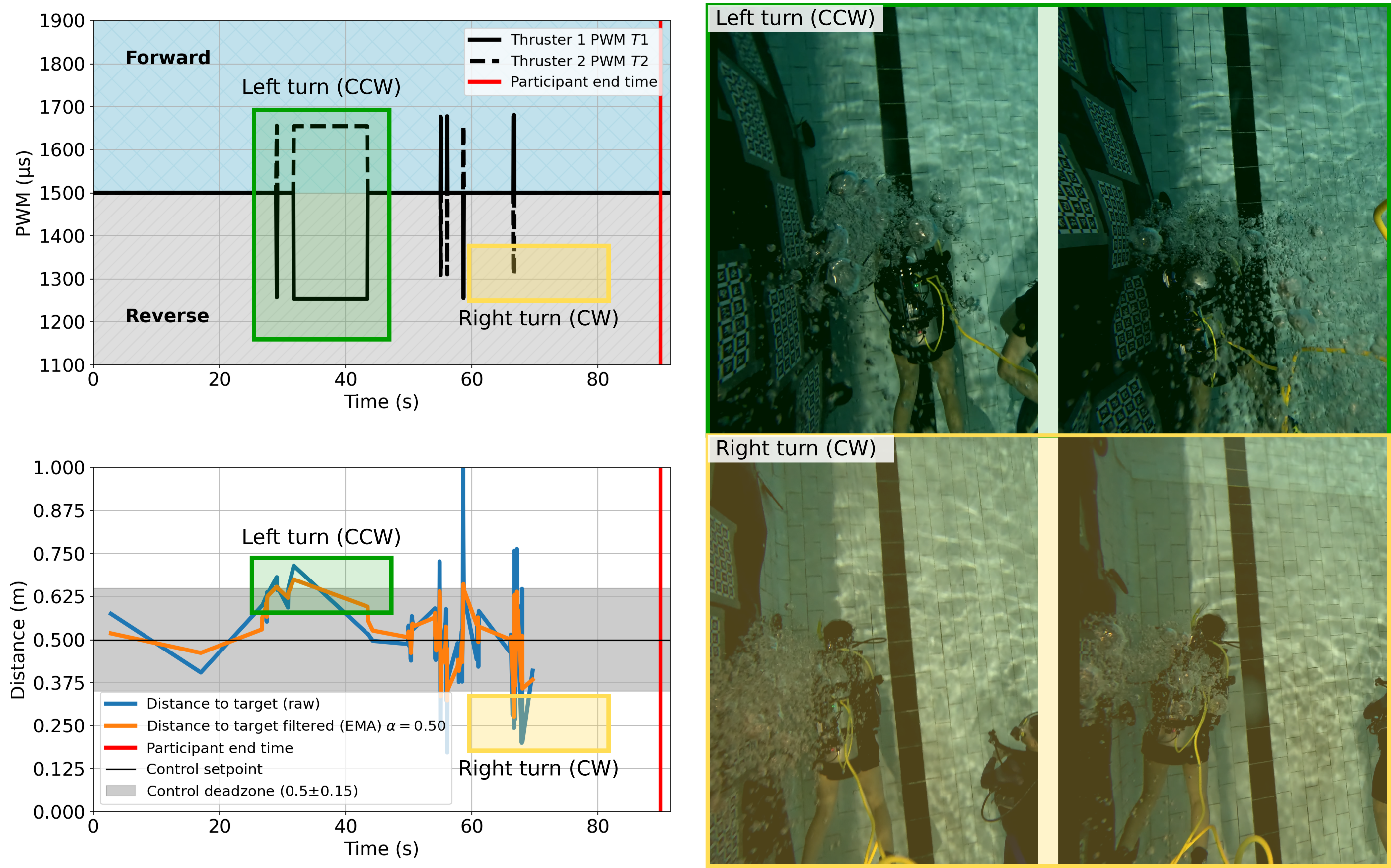}
    \caption{Simulated distance-maintaining experiment showing the correct turn directions based on distance for participant P2. The top plot shows the PWM curves for $T1$ and $T2$. The bottom plot shows the distance to the target for both the raw data and the filtered distance data. The red line shows the participant's end time for the trials, taking $89.89$~s to complete the wall distance. The green region and accompanying aerial perspective show the left turn. The diver responds, turning left. The yellow region shows a right turn, and the diver moves away from the wall.}
    \label{fig:ch6_radmcs_sakshi_trial3}
    % \vspace{-3mm}
\end{figure}

% Add back for journal $DEBUG$
% We show results from two separate data runs with the same participant in Fig.~\ref{fig:ch6_radmcs_sakshi_trial3} and Fig.~\ref{fig:ch6_radmcs_sakshi_trial2}. In each figure, 
We show results for participant P2 in Fig.~\ref{fig:ch6_radmcs_sakshi_trial3}. We plot the PWM versus time curve and the distance versus time curve, all extracted from the robot's data onboard. In each figure, we show the PWM plot for $T1$ and $T2$ with respect to time, capturing the duration of the experimental run. The labeling is the same as in the TS plots. A solid black line denotes thruster 1 $T1$, while a dotted black line denotes thruster 2 $T2$. A dotted line on top (in the forward, blue region) and a solid line on the bottom (in the reverse, gray region) means turn left or yaw counterclockwise. Conversely, a solid line on the top (in the forward, blue region) and a dotted line on the bottom (in the reverse, gray region) means turn right or yaw clockwise. We plot a shaded overlay region for each turn maneuver. A green semi-transparent region means the robot indicated to the diver to turn left, while a yellow semi-transparent region means the robot indicated turn right. On the right hand side of the figure we plot the aerial view for each turn maneuver. We also plot the participant's end time in each case as a solid red line, near the end of each plot. 
% Some data exceeds this boundary, since there was a time delay between the true end time of each participant and when the bench operator terminated the control algorithm. 

Fig.~\ref{fig:ch6_radmcs_sakshi_trial3} shows when the diver exceeds the deadzone boundary (plotted as a gray shaded region in the distance plot on the bottom left plot of the figure) at $t=22$~s, the robot indicates to the diver to turn back toward the wall, since the diver's distance has exceeded the setpoint threshold and deadzone boundary. At $t=68$~s, the opposite happens. The diver swims too close to the wall, with a distance that exceeds the control deadzone boundary on the lower bounds. The robot, in turn, indicates the diver should turn right, away from the wall. 

\begin{figure}[htbp]
\centering
    \includegraphics[width=1.0\linewidth]{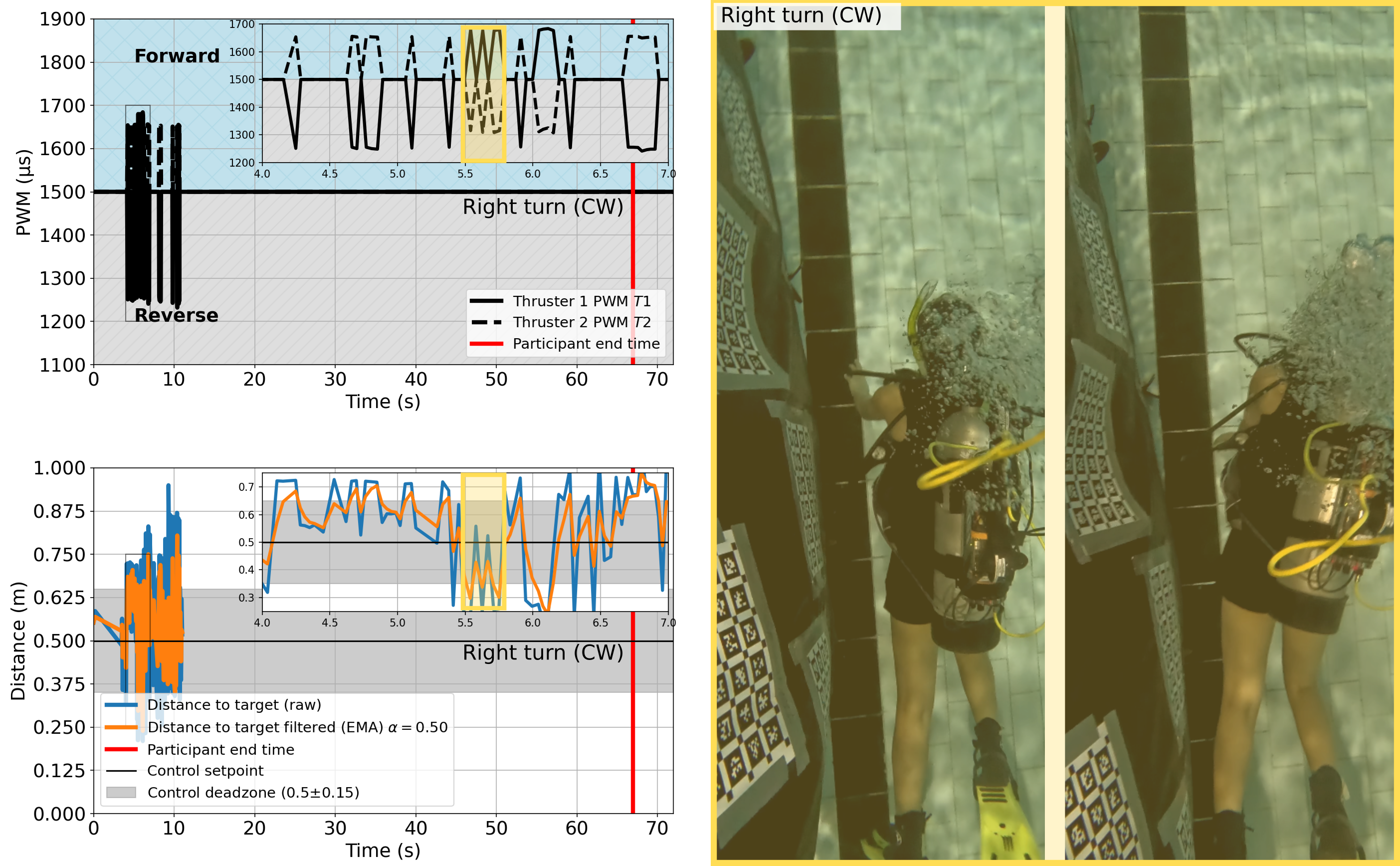}
    \caption{Simulated distance-maintaining experiment showing the correct turn directions based on distance for participant P2. 
    % The top plot shows the PWM curves for $T1$ and $T2$. The bottom plot shows the distance to the target for both the raw data and the filtered distance data. The red line shows the participant's end time for the trials, taking $66.95$~s to complete the wall distance. The yellow region and accompanying aerial perspective show the right turn. The inset image shows the plots from $t=4$~s to $t=7$~s. After this, the distance estimation fails to capture reliable distance data.
    }
    \label{fig:ch6_radmcs_sakshi_trial2}
    % \vspace{-4mm}
\end{figure}

Fig.~\ref{fig:ch6_radmcs_sakshi_trial2} shows that the robot failed to capture distance information past approximately $t=12$~s. We show an inset plot for both the PWM and distance plots from $t=4$~s to $t=7~$s. The inset plot reveals fluctuating PWM signals, from $t=4$~s to $t=5.5$~s the robot indicates to the diver to turn left, toward the wall. However, in the aerial view (not shown) the diver does not respond to this indication. However, at $t=5~$s the robot determines the diver is too close to wall and directs the diver to turn right, away from the wall.

\begin{figure}[htbp]
\centering
    \includegraphics[width=1.0\linewidth]{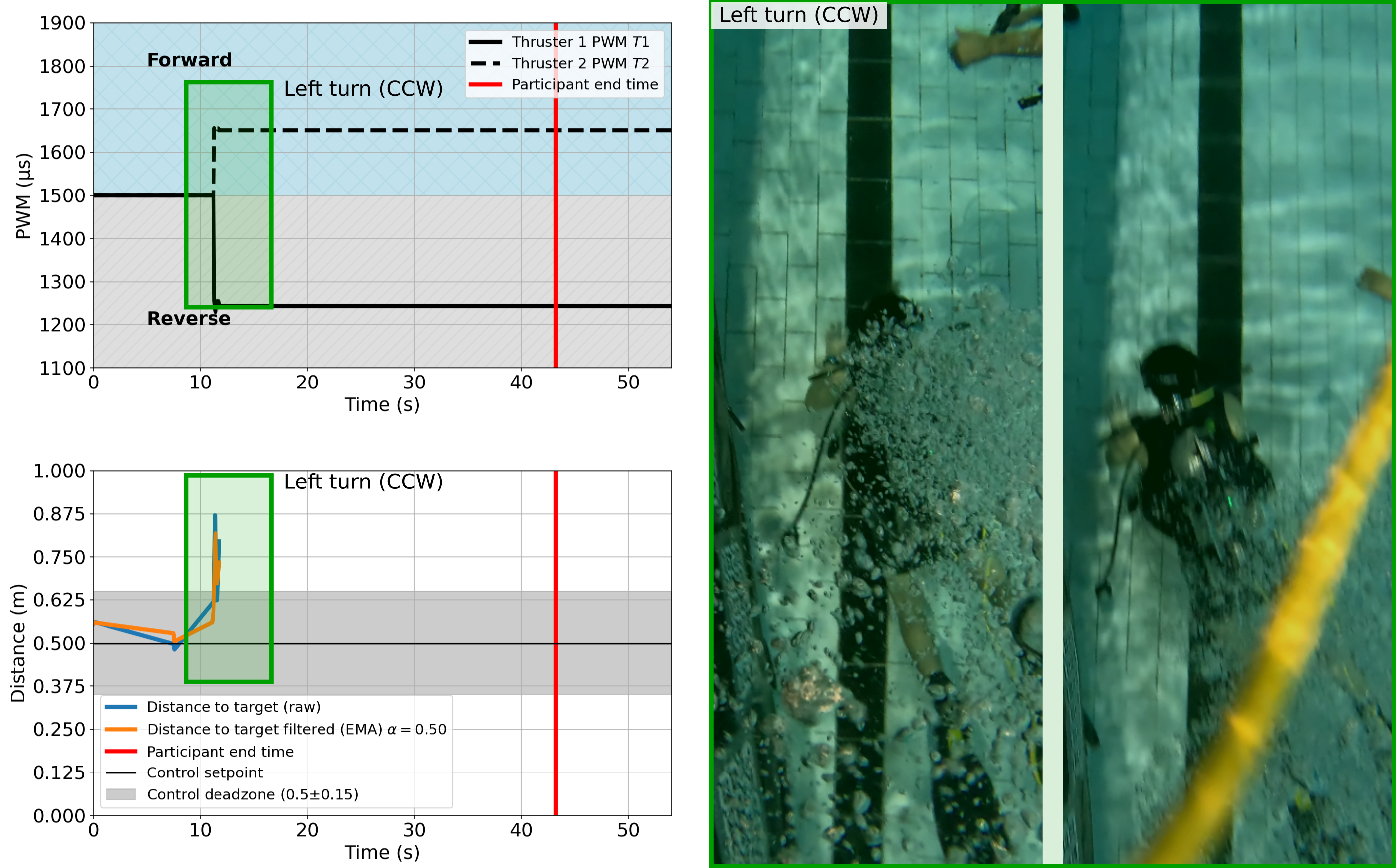}
    \caption{Simulated distance-maintaining experiment showing the correct turn directions based on distance for participant P8. The top plot shows the PWM curves for $T1$ and $T2$. The bottom plot shows the distance to the target for both the raw data and the filtered distance data. The red line shows the participant's end time for the trials, taking $43.28$~s to complete the wall distance. The green region and accompanying aerial perspective show the left turn. Unfortunately, because the robot lost the distance estimation at approximately $t=12~$s of run time and never recovered distance, the filtered distance to target control signal never updated, leaving the robot in a high turn left thrust state for the duration of the data run. }
    \label{fig:ch6_radmcs_demetri_trial2}
    % \vspace{-2mm}
\end{figure}

Fig.~\ref{fig:ch6_radmcs_demetri_trial2} shows that the robot failed to capture distance information past approximately $t=12$~s. Since the robot lost the filtered distance estimate while the diver was beyond the deadzone, the control algorithm never updated, leaving the control signal in a high turn left thrust state for the remainder of the data run. 
\begin{figure}[htbp]
\centering
    \includegraphics[width=1.0\columnwidth]{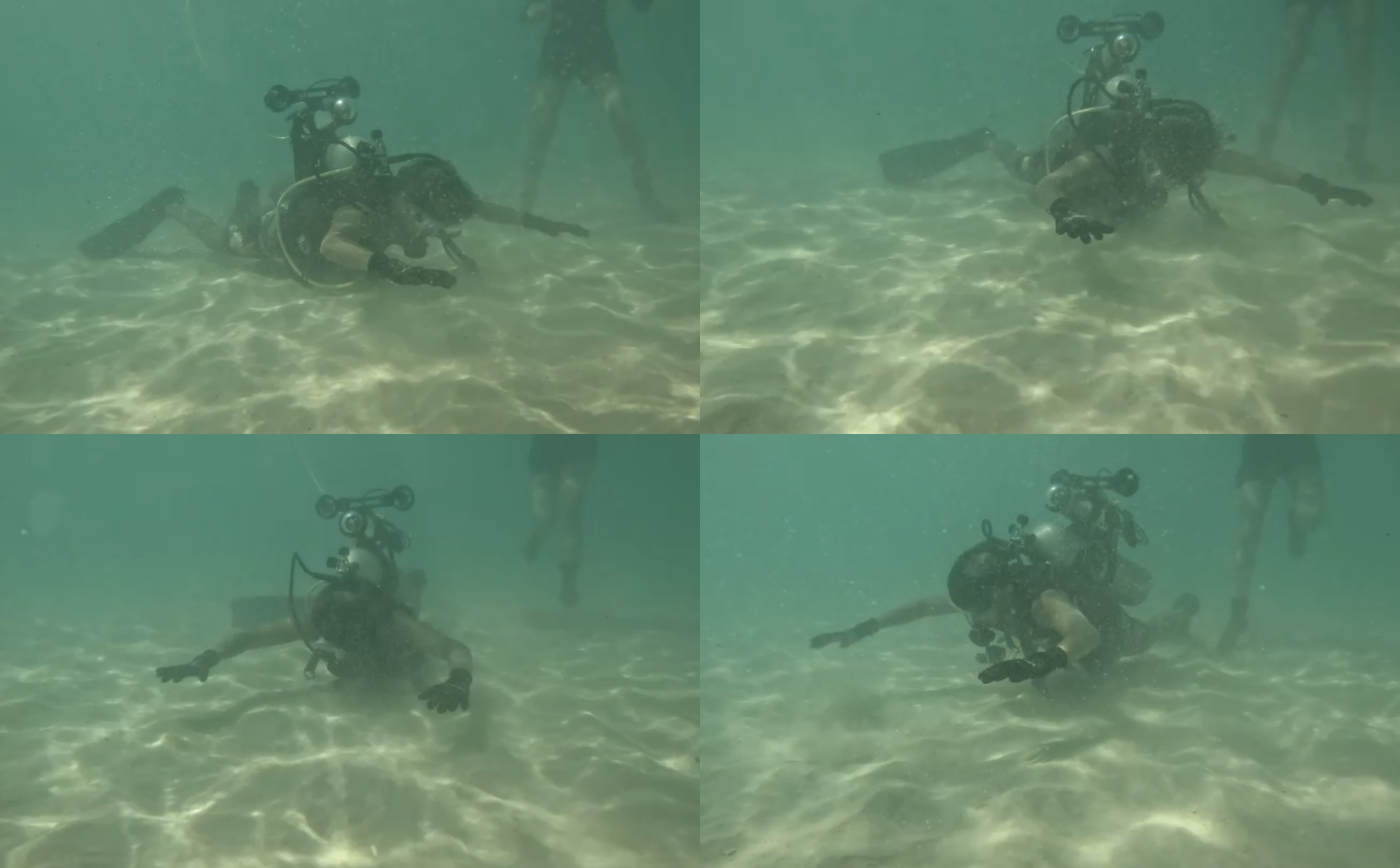}
    \caption{Still frames collected during the open-water ocean TS test for the $325$~mm transverse and longitudinal thruster configurations. During the ``descending'' portion of the test, the thruster force-feedback decreases in intensity, rotating the diver in the figure until they stop moving.}
    \label{fig:radmcs_threshold_ocean_stills}
    % \vspace{-2mm}
\end{figure}

Overall, the single-sided distance-maintaining test reveals that the RADMCS robot can be used for movement assistance when there is a stable and reliable distance estimate.

% \subsection{Open-water Threshold Sensitivity Testing}
\textbf{Open-water Threshold Sensitivity Testing}. 
We conducted two sets of tests in the ocean. First, we completed TS testing for two participants in two thruster configurations: $325$~mm longitudinal and transverse. Second, we performed a maneuverability test to gauge how well the robot assists divers moving through coral reef spaces using the $325$ longitudinal thruster configuration. 
Fig.~\ref{fig:radmcs_threshold_ocean_stills} shows the experimental setup for threshold sensitivity testing in the ocean environment. A safety diver monitored participant safety during the test. 

\begin{figure}[thb]
    \centering  
    
    \subfloat[CW rotation/longitudinal thruster configuration.]{%
        \label{fig:ocean_200mm_cw_longitudinal}%
        \includegraphics[width=0.60\columnwidth]{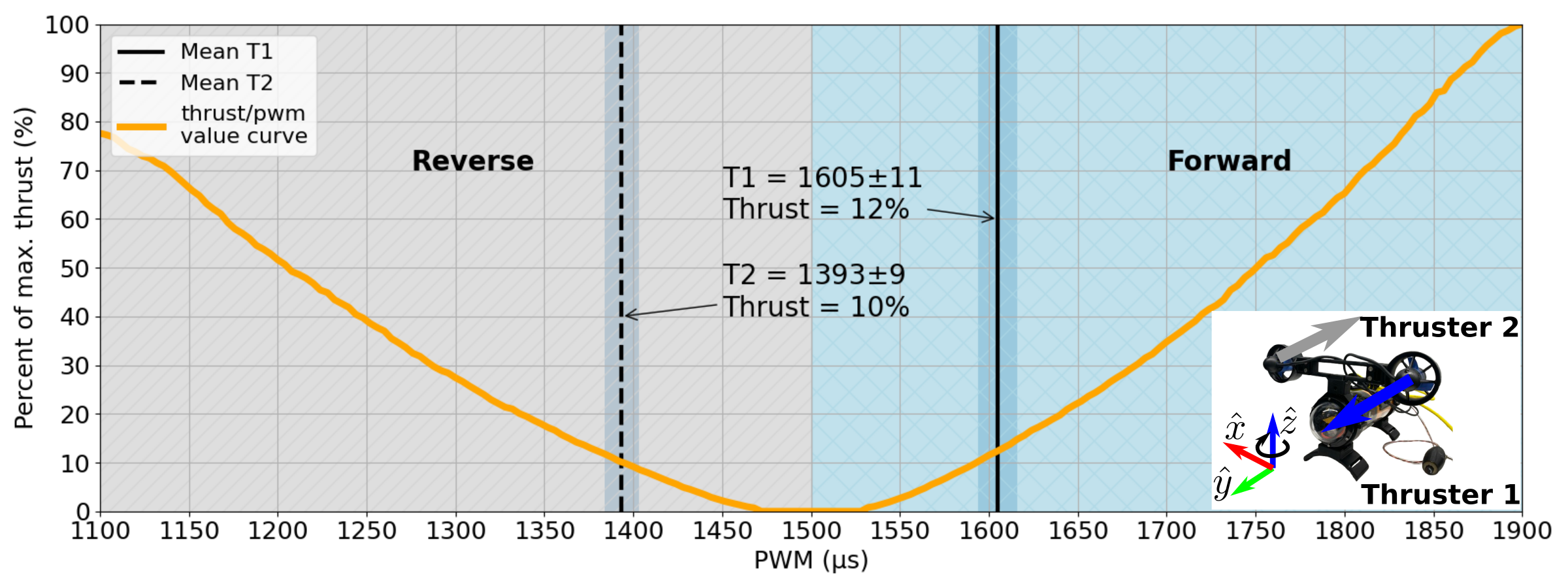}%
    }
    
    \subfloat[CCW rotation/longitudinal thruster configuration.]{%
        \label{fig:ocean_200mm_ccw_longitudinal}%
        \includegraphics[width=0.60\columnwidth]{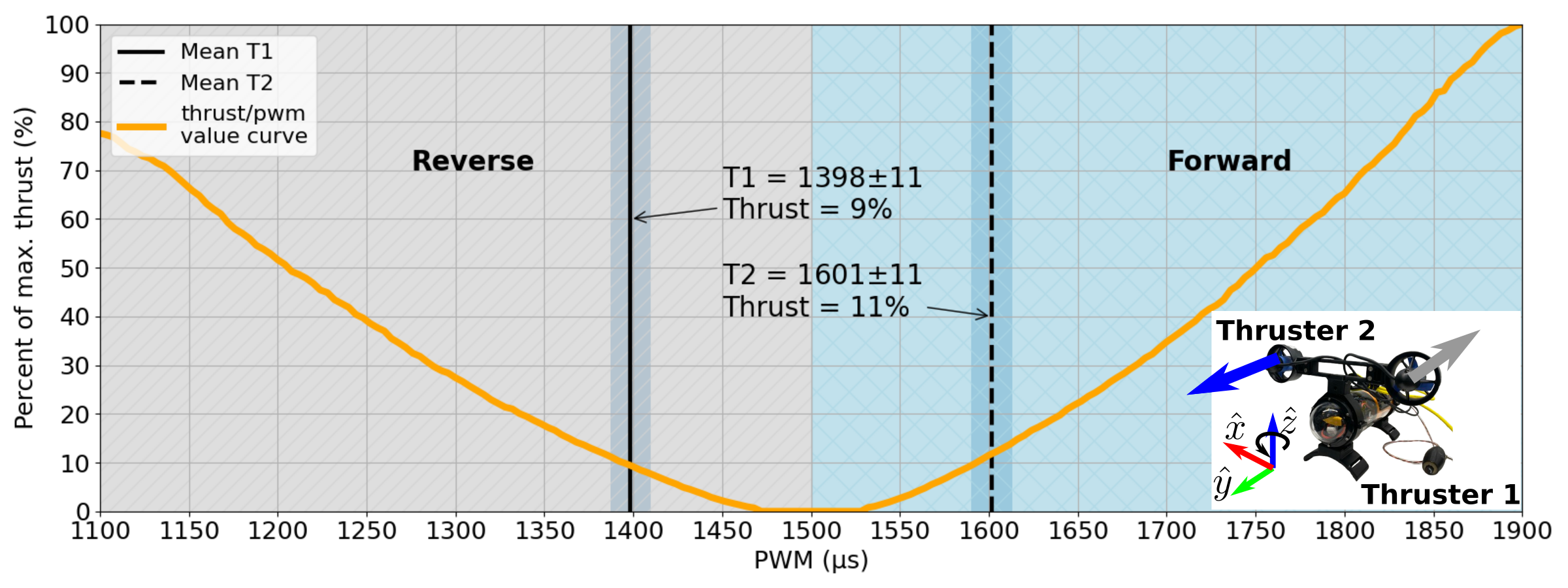}%
    }

    \subfloat[CW rotation/transverse thruster configuration.]{%
        \label{fig:ocean_200mm_cw_transverse}%
        \includegraphics[width=0.60\columnwidth]{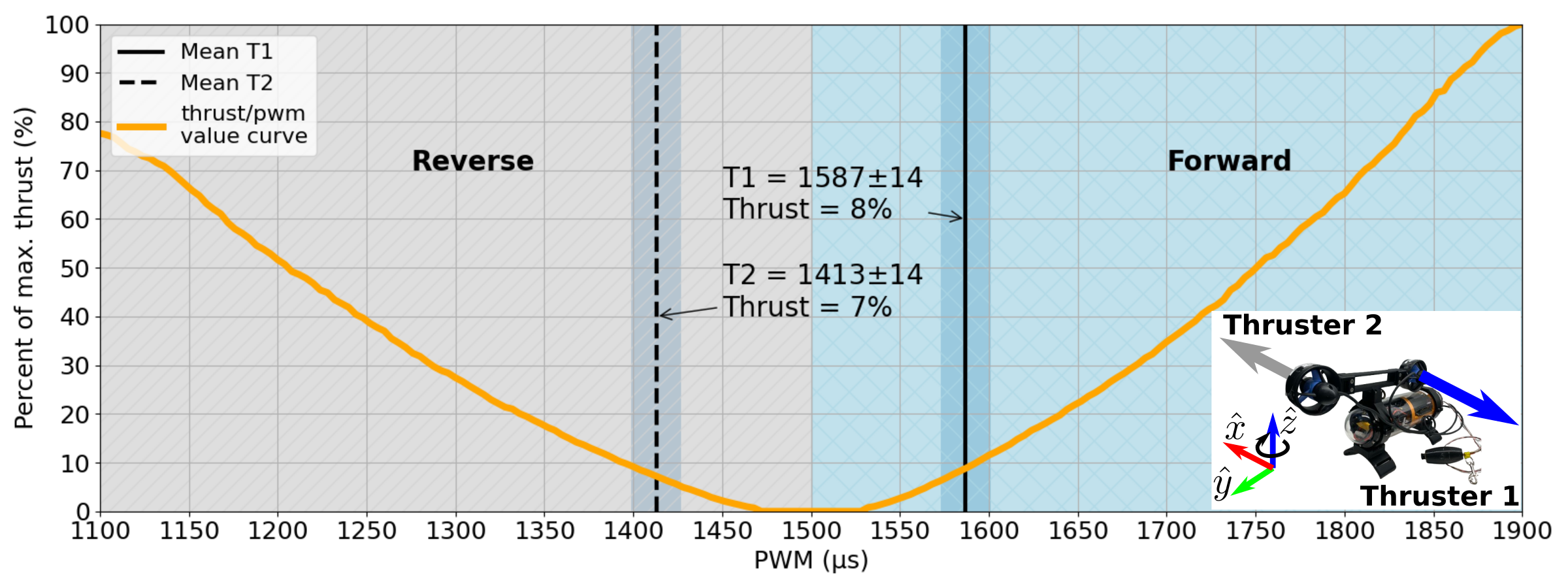}%
    }
    
    \subfloat[CCW rotation/transverse thruster configuration.]{%
        \label{fig:ocean_200mm_ccw_transverse}%
        \includegraphics[width=0.60\columnwidth]{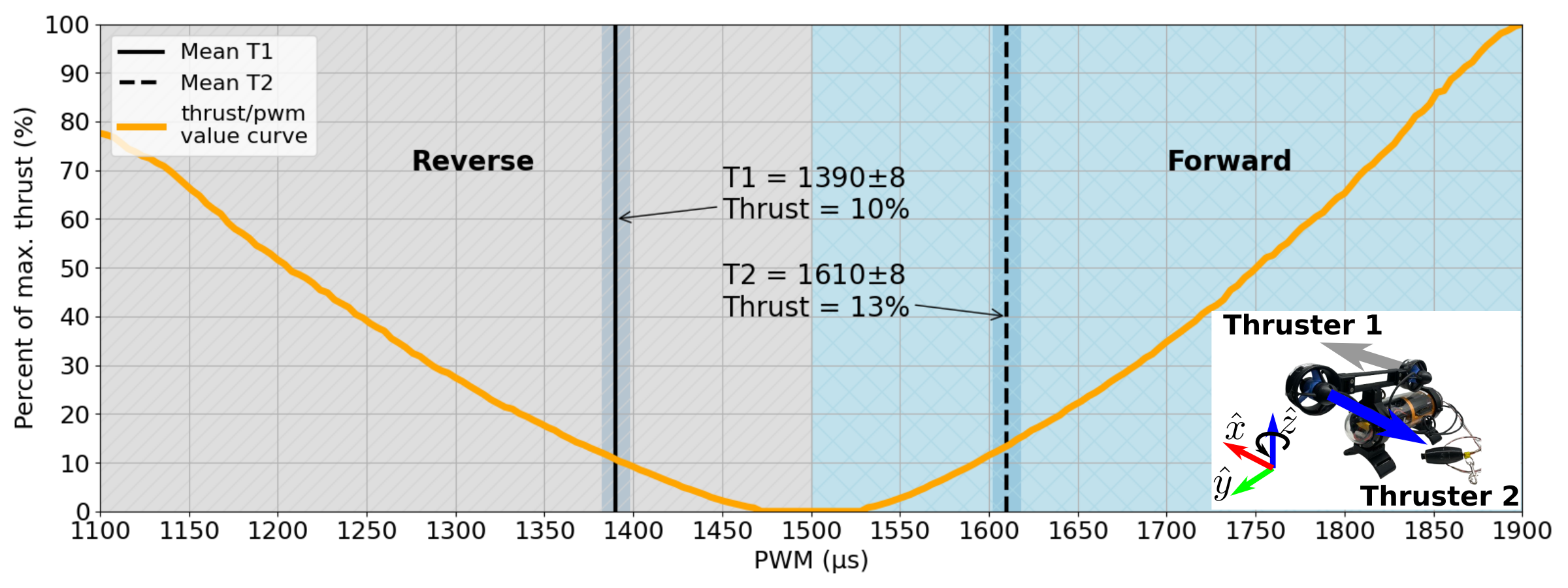}%
    }    
    
    \caption{TS for ($M=2$) participants P1 and P3 for CW and CCW rotations in both the $325$~mm longitudinal and transverse thruster configurations in the ocean environment.}
    \label{fig:radmcs_ocean_325mm}
    % \vspace{-3mm}
\end{figure}

Fig.~\ref{fig:radmcs_ocean_325mm} shows the results for the sensitivity test in the ocean. A tether connection was used from the shoreline to the diver to perform the method of limits test for TS. This was done for two participants in each thruster configuration. The participants deployed $5$~m from shore, in a depth of $1.5$~m. We hypothesized that the currents in the ocean environment, particularly near the shore with moderate wave action (wave heights fluctuated between $0.25$--$0.5$~m), would increase the sensitivity limit, since the diver should require more force-feedback in the presence of external disturbances. However, the data shows that again $10$\% of maximum thrust is required to produce perceptible motion in both configurations. This indicates that either the divers felt external disturbances as robotic feedback and could not differentiate between robotic directed force-feedback and subtle environmental factors, or the robot tightly couples to the physical sensing capabilities of the human, regardless of the external disturbances which act along different force directions. Without additional testing, we cannot conclusively prove either hypothesis. We can only conclude that the configuration enables sensitivity levels at $10$\% of maximum thrust available.

\begin{figure}[htbp]
% \vspace{2mm}
\centering
    \includegraphics[width=1.0\columnwidth]{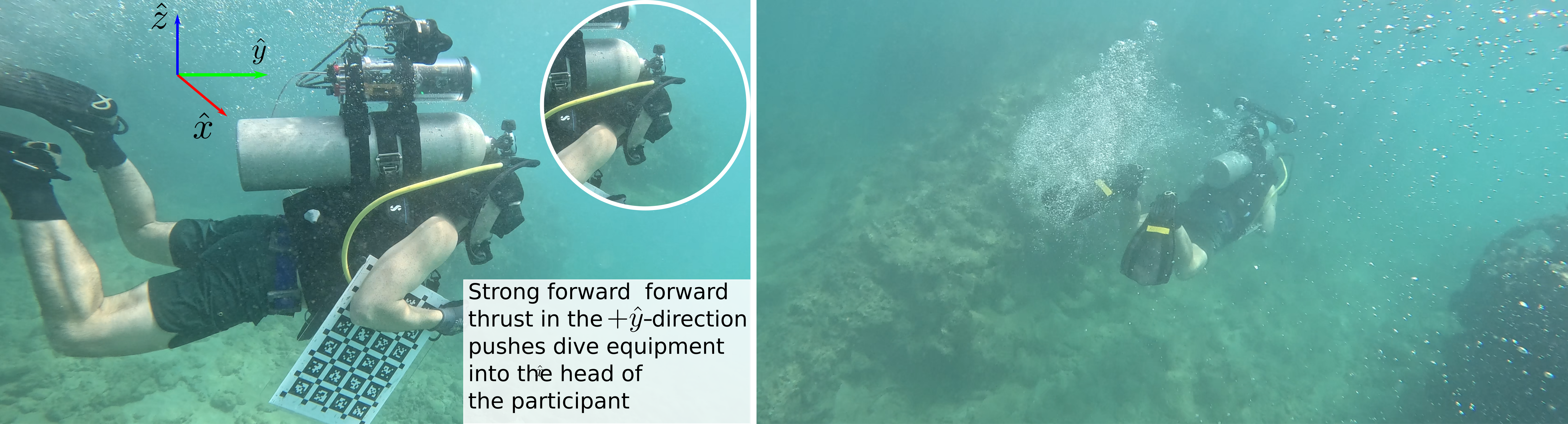}
    \caption{Still frames from the form, fit, and function qualitative test in the ocean environment.}
    \label{fig:radmcs_ocean_form_fit_function}
    % \vspace{-3mm}
\end{figure}
% \subsection{Open-water Form, Fit, and Function Qualitative Experiments}
\textbf{Open-water Form, Fit, and Function Qualitative Experiments}.
We then conducted an experiment to gauge form, fit, and function in the open-water environment. We did this for one participant, P8, given logistical complexity. The participant was outfitted with the RADMCS system in the $325$~mm longitudinal thruster configuration. The thrust value for each thruster $T1$ and $T2$ was set to $1850~\mu$s, or $80$\% of maximum thrust according to the manufacturer's nonlinear thrust vs. PWM curve (see the gold curve in Fig.~\ref{fig:radmcs_pool_225mm}). The diver was instructed to turn the system on at a depth of $4$~m, use the robot to navigate the coral reef, and qualitatively gauge how well the robot (acting purely as a movement device) can assist in swimming. Fig.~\ref{fig:radmcs_ocean_form_fit_function} shows the experimental setup.

The diver reported that the device moved out of place during the experiment, creating a painful pinch point at the diver's head. This is because the robot's forward thrust in the $+\hat{y}$-direction coupled into the diver's equipment, shifting it forward. This displacement moved the regulator and hoses too far forward for extended use applications. However, accounting for small swimming requirements to adjust body position in the water column, the diver reported they could rely on the device for swimming assistance without the need for supplemental movement swimming.

\section{Discussion \& Limitations}
\label{ch6:radmcs_discussion}
We conducted significant underwater testing of the RADMCS robotic system, including TS tests in all four thruster configurations; simulated wall-following experiments, and open-water form, fit, and function experiments to gauge the qualitative and practical limitations of the current design. We determined that $10$\% of maximum thrust is required to produce a  perceptible effect from the robot to the human, and we also determined that when stable distance estimates are present, the robot can correctly indicate navigational information to the human using thruster value control. Given the logistical complexity of human studies in underwater robotics, we were limited to a total of eight study participants. More comprehensive testing is warranted to explore aspects of the system such as deep learning-based depth estimation, \eg the work of \citet{yang2025seasplat} for underwater three-dimensional reconstruction. The relative simplicity of the linear control algorithm with saturation creates opportunity for complex model-based control dynamics which not only take into account distance as a control signal but also the human's feedback to the robot \citep{molton1998stereo,liao2020human}.
\vspace{-1mm}
\section{Conclusion}
\label{ch6:radmcs_conclusion}
% We have introduced a novel and early-stage robot for diver movement assistance underwater, leveraging horizontal-facing cameras that can perform depth estimation coupled to haptic feedback to keep divers away from potential hazards. Our preliminary results indicate we are making progress toward creating a system that is robust and will hopefully provide a foundation for continued innovation in this research area. Continued testing will explore the bounds and applicability of the current design. Specifically, we will leverage the detection threshold results to tailor the haptic feedback mechanism to help a cohort of divers navigate submerged navigation courses.

We present RADMCS, a novel wearable robotic system designed to assist divers in maintaining safe standoff distances in confined or hazardous underwater environments. Although further testing is warranted for TS in open-water environments with additional study participants, the system as proposed shows promise for foundational work in robotic-assisted navigation underwater. We will expand testing to include closed-water and open-water navigation tasks, evaluate the generality of threshold values across a broader diver population, and refine control strategies for more complex behaviors such as simulated confined-space traversal. These efforts will help establish RADMCS as a robust platform for physical underwater human-robot interaction and hopefully garner significant interest for broader community investment in safety-critical wearable robots underwater.

% \clearpage
\begingroup
  \renewcommand{\UrlFont}{\ttfamily\tiny}
  \bibliographystyle{ieeetr}
  \bibliography{ch6_radmcs_bibliography}

@article{byrd1997_underwater_cave_fatalities,
  title={{Underwater cave diving fatalities in Florida: A review and analysis}},
  author={Byrd, JH and Hamilton, WF},
  journal={Journal of Forensic Sciences},
  volume={42},
  number={5},
  pages={807--811},
  year={1997},
  publisher={American Society of Mechanical Engineers Digital Collection}
}

@misc{ads_cost,
    author={Nick Wright},
    year={2014},
    note={Accessed on 2026-09-15},
    title={Into the abyss},
    howpublished={\url{https://www.ffjournal.net/industry-news/13-special-report/175-extreme-fabricating/12616-into-the-abyss}},
}

@book{bradski2008learning,
  title={Learning {OpenCV}: Computer vision with the {OpenCV} library},
  author={Bradski, Gary and Kaehler, Adrian},
  year={2008},
  doi={10.5555/2666676},
  publisher={ O'Reilly Media, Inc.}}

@inproceedings{oth2013rolling,
  title={{Rolling shutter camera calibration}},
  author={Oth, Luc and Furgale, Paul and Kneip, Laurent and Siegwart, Roland},
  booktitle={Proceedings of the IEEE/CVF Conference on Computer Vision and Pattern Recognition (CVPR)},
  pages={1360--1367},
  year={2013}
}

@misc{dsend,
    author={{Johns Hopkins Applied Physics Laboratory}},
    howpublished={\url{https://www.jhuapl.edu/work/projects-and-missions/dsend}},
    note={Accessed: 2026-09-15},
}

@article{ros2,
    author = {Steven Macenski  and Tully Foote  and Brian Gerkey  and Chris Lalancette  and William Woodall },
    title = {{Robot Operating System 2: Design, architecture, and uses in the wild}},
    journal = {Science Robotics},
    volume = {7},
    number = {66},
    pages = {eabm6074},
    year = {2022},
    doi = {10.1126/scirobotics.abm6074},
    URL = {https://www.science.org/doi/abs/10.1126/scirobotics.abm6074},
}

@ARTICLE{jones2013_haptic_threshold,
  author={Jones, Lynette A. and Tan, Hong Z.},
  journal={IEEE Transactions on Haptics}, 
  title={{Application of psychophysical techniques to haptic research}}, 
  year={2013},
  volume={6},
  number={3},
  pages={268--284},
  doi={10.1109/TOH.2012.74}}

@article{marston2006evaluation_spatial,
  title={Evaluation of spatial displays for navigation without sight},
  author={Marston, James R and Loomis, Jack M and Klatzky, Roberta L and Golledge, Reginald G and Smith, Ethan L},
  journal={ACM Transactions on Applied Perception (TAP)},
  volume={3},
  number={2},
  pages={110--124},
  year={2006},
  publisher={ACM New York, NY, USA}
}

@inproceedings{golgouneh2021design_soft_exo_skeleton,
  title={{Design of a hybrid SMA-pneumatic based wearable upper limb exoskeleton}},
  author={Golgouneh, Alireza and Beaudette, Eric and Woelfle, Heidi and Li, Bai and Subash, Niharikha and J. Redhouse, Amanda and Jones, Mark and Martin, Tom and A. Lobo, Michele and Holschuh, Brad and others},
  booktitle={Proceedings of the ACM International Symposium on Wearable Computers},
  pages={179--183},
  year={2021}
}

@article{subash2025haptic_haptics_vibro_compression,
  title={Haptic perception interaction effects between vibrotactile and compressive stimuli in wearable devices},
  author={Subash, Niharikha and Holschuh, Brad},
  journal={The Design Journal},
  pages={1--20},
  year={2025},
  publisher={Taylor \& Francis}
}

@inproceedings{prabhu2020vibrosleeve_sleeve_mounted,
  title={{VibroSleeve: A wearable vibro-tactile feedback device for arm guidance}},
  author={Prabhu, Deepa and Hasan, Muhammad Mehedi and Wise, Lisa and MacMahon, Clare and McCarthy, Chris},
  booktitle={Proceedings of the Annual International Conference of the IEEE Engineering in Medicine \& Biology Society (EMBC)},
  pages={4909--4912},
  year={2020},
  organization={IEEE}
}

@article{loomis2005personal_guidance,
  title={{Personal guidance system for people with visual impairment: A comparison of spatial displays for route guidance}},
  author={Loomis, Jack M and Marston, James R and Golledge, Reginald G and Klatzky, Roberta L},
  journal={Journal of Visual Impairment \& Blindness},
  volume={99},
  number={4},
  pages={219--232},
  year={2005},
  publisher={SAGE Publications Sage CA: Los Angeles, CA}
}

@inproceedings{pielot2010pocketnavigator_haptics,
  title={{PocketNavigator: Vibro-tactile waypoint navigation for everyday mobile devices}},
  author={Pielot, Martin and Poppinga, Benjamin and Boll, Susanne},
  booktitle={{Proceedings of the International Conference on Human Computer Interaction with Mobile Devices and Services (MobileHCI)}},
  pages={423--426},
  year={2010}
}

@inproceedings{tsukada2004activebelt_wearable_guidance,
  title={{Activebelt: Belt-type wearable tactile display for directional navigation}},
  author={Tsukada, Koji and Yasumura, Michiaki},
  booktitle={Proceedings of the International Conference on Ubiquitous Computing},
  pages={384--399},
  year={2004},
  organization={Springer}
}

@article{oncticua2024_eod_divers,
  title={The role of explosive ordnance disposal divers in eliminating the danger generated by naval mines},
  author={On{\c{t}}ic{\u{a}}, Adrian and G{\"o}rde{\c{s}}, Alexandra},
  journal={Land Forces Academy Review},
  volume={29},
  number={2},
  pages={219--226},
  year={2024},
  publisher={Nicolae Balcescu}
}

@article{beckwith2012_diving_impact_on_petroleum,
  title={{Under pressure: Diving's impact on the offshore petroleum industry}},
  author={Beckwith, Robin},
  journal={Journal of Petroleum Technology},
  volume={64},
  number={04},
  pages={52--60},
  year={2012},
  publisher={SPE}
}

@article{lin2021_touching_coral_reef,
  title={{Close encounters of the worst kind: reforms needed to curb coral reef damage by recreational divers}},
  author={Lin, Bing},
  journal={Coral Reefs},
  volume={40},
  number={5},
  pages={1429--1435},
  year={2021},
  publisher={Springer}
}

@book{adci_standards,
    author = {},
    title = {International Consensus Standards for Commercial Diving and Underwater Operations},
    publisher = {Association of Diving Contractors International, Inc.},
    year = {2024},
    edition = {6},
}

@misc{blueRobotics_thrusters, 
	title = {T200 thrusters},
	author={{Blue Robotics}},
    year = {2022},
	howpublished = {\url{https://bluerobotics.com/store/thrusters/t100-t200-thrusters/t200-thruster-r2-rp/}},
    note = {Accessed: 2026-09-15},
}

@misc{blueRobotics_enclosure,
    title={Watertight Enclosure Tubes},
    howpublished={\url{https://bluerobotics.com/store/watertight-enclosures/locking-series/wte-locking-tube-r1-vp/}},
    note={Accessed: 2026-09-15},
    author={{Blue Robotics}},
    year={2022},
}

@misc{bluetrail, 
	title = {{Cobalt Series Bulkhead Connector}},
	author={{Blue Trail Engineering}},
    year = {2026},
	howpublished = {\url{https://www.bluetrailengineering.com}},
    note = {Accessed: 2026-04-19},
}

@article{liao2020human,
  title={Human navigation using phantom tactile sensation based vibrotactile feedback},
  author={Liao, Zhenyu and Luces, Jose V Salazar and Hirata, Yasuhisa},
  journal={IEEE Robotics and Automation Letters},
  volume={5},
  number={4},
  pages={5732--5739},
  year={2020},
  publisher={IEEE}
}

@article{molton1998stereo,
  title={A stereo vision-based aid for the visually impaired},
  author={Molton, Nicholas and Se, Stephen and Brady, JM and Lee, David and Probert, Penny},
  journal={Image and Vision Computing},
  volume={16},
  number={4},
  pages={251--263},
  year={1998},
  publisher={Elsevier}
}

@inproceedings{yang2025seasplat,
  title={{SeaSplat: Representing underwater scenes with 3D gaussian splatting and a physically grounded image formation model}},
  author={Yang, Daniel and Leonard, John J and Girdhar, Yogesh},
  booktitle={Proceedings of the IEEE International Conference on Robotics and Automation (ICRA)},
  pages={7632--7638},
  year={2025},
  organization={IEEE}
}

@article{iliffe2001_cave_diving,
  title={Scientific cave diving},
  author={Iliffe, Thomas M and Bowen, Curt},
  journal={Marine Technology Society Journal},
  volume={35},
  number={2},
  pages={36--41},
  year={2001},
  publisher={Marine Technology Society}
}

@book{exley1986_basic_cave_diving,
  title={{Basic cave diving: A blueprint for survival}},
  author={Exley, Sheck},
  year={1986},
  publisher={Cave Diving Section of the National Speleological Society}
}

@inproceedings{abdullah2024_Cave_Seg,
  title={{Caveseg: Deep semantic segmentation and scene parsing for autonomous underwater cave exploration}},
  author={Abdullah, Adnan and Barua, Titon and Tibbetts, Reagan and Chen, Zijie and Islam, Md Jahidul and Rekleitis, Ioannis},
  booktitle={Proceedings of the IEEE International Conference on Robotics and Automation (ICRA)},
  pages={3781--3788},
  year={2024},
  organization={IEEE}
}

@article{Li2022UnderwaterSurvey,
  title={{Wearable robots for human underwater movement ability enhancement: A survey}},
  author={Li, X. and Zhang, Y. and Chen, X.},
  journal={IEEE/CAA Journal of Automatica Sinica},
  year={2022},
  volume={9},
  number={4},
  pages={777--793},
  doi={10.1109/JAS.2022.1234567}
}

@article{wang2025system_soft_exoskeleton,
  title={{System design of a soft underwater exosuit to reduce metabolic cost across multiple aquatic movements during diving}},
  author={Wang, Xiangyang and Chen, Chunjie and Sun, Jianquan and Du, Sida and Ma, Yue and Wu, Xinyu},
  journal={IEEE Transactions on Robotics},
  year={2025},
  publisher={IEEE}
}

@article{wang2020underwater_soft_exoskeleton,
  title={An underwater lower-extremity soft exoskeleton for breaststroke assistance},
  author={Wang, Qining and Zhou, Zhihao and Zhang, Zhendong and Lou, Yu and Zhou, Yalei and Zhang, Shichang and Chen, Wanwen and Mao, Chengqiang and Wang, Zilu and Lou, Wenjie and others},
  journal={IEEE Transactions on Medical Robotics and Bionics},
  volume={2},
  number={3},
  pages={447--462},
  year={2020},
  publisher={IEEE}
}

@article{zeng2016control,
  title={Control of hydraulic drive system of underwater exoskeleton robot},
  author={Zeng, D},
  journal={MS thesis, Sch. Aeronaut. Astronaut},
  year={2016}
}

@article{erp2005_waypoint_haptics,
  title={Waypoint navigation with a vibrotactile waist belt},
  author={Erp, Jan BF Van and Veen, Hendrik AHC Van and Jansen, Chris and Dobbins, Trevor},
  journal={ACM Transactions on Applied Perception (TAP)},
  volume={2},
  number={2},
  pages={106--117},
  year={2005},
  publisher={ACM New York, NY, USA}
}

@misc{NuytcoExosuitSpec,
  title={{Exosuit atmospheric diving system (ADS) technical specifications}},
  author={{Nuytco Research Ltd.}},
  year={2021},
  url={https://nuytco.com/exosuit},
  note={Accessed 2025-09-30}
}

@inproceedings{Neuhaus2004SwimmingExo_rigid,
  title={{Concept designs for underwater swimming exoskeletons}},
  author={Neuhaus, P. and Kazerooni, H.},
  booktitle={Proceedings of the IEEE International Conference on Robotics and Automation (ICRA)},
  year={2004},
  pages={1234--1240},
  doi={10.1109/ICRA.2004.1234567}
}

@Article{ma15082841,
  AUTHOR = {Pje{\v{s}}{\v{c}}i{\'c}-{\v{S}}{\'c}epanovi{\'c}, Jelena and Vastag, Gy{\"o}ngyi and Ivo{\v{s}}evi{\'c}, {\v{S}}piro and Kova{\v{c}}, Nata{\v{s}}a and Rudolf, Rebeka},
  TITLE = {{Corrosion of NiTiDiscs in different seawater environments}},
  JOURNAL = {Materials},
  VOLUME = {15},
  YEAR = {2022},
  NUMBER = {8},
  ARTICLE-NUMBER = {2841},
  PubMedID = {35454534},
  ISSN = {1996--1944},
  DOI = {10.3390/ma15082841}
}

@article{YANG2024117005,
title = {Design and analysis of a shape memory alloy buoyancy actuator},
journal = {Ocean Engineering},
volume = {296},
pages = {117005},
year = {2024},
issn = {0029-8018},
doi = {10.1016/j.oceaneng.2024.117005},
author = {Shaoqiong Yang and Lintao Jia and Anqi Sun and Xu Wan and Wei Ma and Wendong Niu and Ming Yang and Tongshuai Sun},
}
\endgroup
\end{document}